\documentclass[sn-basic,onecolumn]{sn-jnl}% Basic Springer Nature Reference Style/Chemistry Reference Style
\usepackage{mathtools}
\usepackage{amsmath,amssymb,amsfonts,bm}
\usepackage{graphicx}
\graphicspath{{figs/}}
\usepackage{enumitem}
\usepackage{subcaption}
\usepackage{float,placeins}
\definecolor{pumpkin}{rgb}{1.0, 0.46, 0.09}

\usepackage{chngcntr}
\counterwithin{figure}{section}
\usepackage{longtable} % Add this in the preamble
\usepackage{dblfloatfix}
\usepackage{fix-cm}

\usepackage{hyperref}
\hypersetup{
    colorlinks=true,
    linkcolor=blue,
    filecolor=magenta,      
    urlcolor=blue,
}
\usepackage[para,online,flushleft]{threeparttable}
\usepackage{multirow}

\jyear{2022}%

\theoremstyle{thmstyleone}%
\theoremstyle{thmstyletwo}%

\theoremstyle{thmstylethree}%

\usepackage{csquotes}
\usepackage{siunitx}

\begin{document}

\title[Ensemble Modelling of Groundwater Metal Pollution]{Smart Ensemble Learning Framework for Predicting Groundwater Heavy Metal Pollution} 


\author[1]{\fnm{T.} \sur{Ansah-Narh}}
\author*[1]{\fnm{G. Y.} \sur{Afrifa}}\email{george.afrifa@gaec.gov.gh, +233249243042}    
\author[1]{\fnm{J. B.} \sur{Tandoh}}
\author[1]{\fnm{K.} \sur{Asare}} 
\author[1]{\fnm{M.} \sur{Addi}}
\author[1]{\fnm{K. E.} \sur{Yorke}}
\author[1]{\fnm{D. M. A. } \sur{Akpoley}}
\author[1]{\fnm{K.} \sur{Aidoo}}
\author[1]{\fnm{S. K.} \sur{Fosuhene}}


% \author[1]{\fnm{Kwabina} \sur{Ibrahim}} 

% \author*[1]{\fnm{Yvonne S. A.} \sur{Loh}}\email{sloh@ug.edu.gh}

% \author[1]{\fnm{Patrick A.} \sur{Sakyi}} 

% \author[1]{\fnm{Larry P.} \sur{Chegbeleh}} 

% \author[1]{\fnm{Sandow M.} \sur{Yidana}} 
 \affil[1]{\orgdiv{Ghana Space Science and Technology Institute}, \orgname{Ghana Atomic Energy Commission}, \orgaddress{ \postcode{P. O. Box LG 80}, \state{Accra}, \country{Ghana}}}
% \affil[2]{\orgdiv{Department of Geography}, \orgname{University of Cape Coast}, \orgaddress{\street{street..}, \city{Cape Coast}, \postcode{P.O. Box --}, \country{Ghana}}}
% \affil[3]{\orgdiv{Department of Physics}, \orgname{University of Ghana}, \orgaddress{\street{street..}, \city{Accra}, \postcode{P.O. Box --}, \country{Ghana}}}

% \affil[2]{\orgdiv{Ghana Space Science and Technology Institute}, \orgname{Ghana Atomic Energy Commission}, \orgaddress{ \postcode{P. O. Box LG 80}, \state{Accra}, \country{Ghana}}}

% \affil[3]{\orgdiv{Department}, \orgname{Organization}, \orgaddress{\street{Street}, \city{City}, \postcode{610101}, \state{State}, \country{Country}}}


\abstract{
Groundwater in the Densu Basin is increasingly threatened by heavy metal contamination, yet conventional assessment methods struggle to capture the statistical complexity and spatial heterogeneity of pollution indicators. A critical challenge lies in modelling the Heavy Metal Pollution Index (HPI), which is typically skewed and influenced by correlated contaminants, leading to biased predictions when modelled without transformation. This study develops a predictive framework that integrates response transformations with nested cross-validated ensemble machine learning to address these limitations.
Three transformations; raw, log, and Gaussian copula, were applied to HPI and evaluated across six learners: support vector regression (SVM), $k$-nearest neighbours (k-NN), CART, Elastic Net, kernel ridge regression, and a stacked Lasso ensemble. Diagnostic evaluation showed that raw-scale models produced deceptively high fits, with Elastic Net and the stacked ensemble reporting $R^2$ values close to $1.0$, raising concerns of over-optimism and potential information leakage. The log transformation stabilised variance, improving prediction for SVM ($R^2 = 0.93$, RMSE $= 0.18$) and k-NN ($R^2 = 0.92$, RMSE $= 0.20$), though the performance of Elastic Net deteriorated. The Gaussian copula transformation yielded the most reliable outcomes: the stacked ensemble achieved $R^2 = 0.96$ with RMSE $= 0.19$, while other learners such as SVM ($R^2 = 0.86$, RMSE $= 0.25$) and k-NN ($R^2 = 0.85$, RMSE $= 0.26$) maintained high accuracy. Importantly, copula-based models improved residual behaviour and produced spatially plausible prediction maps, reinforcing their potential for groundwater quality management. Clustering analysis using DBSCAN further revealed the dominance of Fe, followed by Mn, as the primary contributors to HPI, consistent with regional hydrogeochemical processes.
Limitations include the reliance on random rather than spatial cross-validation and the basin-specific nature of the analysis, which may constrain transferability. Future research should explore spatially explicit validation schemes and extend the framework to diverse hydrogeological settings. Overall, the study advances predictive hydrogeochemistry by demonstrating that distribution-aware ensembles, complemented by clustering diagnostics, can provide robust and interpretable assessments of groundwater contamination.
}
 

% \maketitle
%----------------------------------------
% 2. Highlights (on a separate page)
%----------------------------------------
\clearpage
\section*{Highlights}
\begin{enumerate}
    \item  Applied rank-based methods to handle skewed and tied observations.

\item Compared raw, log, and copula transformations for predictive accuracy.

\item DBSCAN clustering revealed Fe and Mn as dominant HPI contributors.

\item Used nested CV to prevent overfitting and ensure unbiased evaluation.

\item Achieved high accuracy ($ \rm R^2 = 0.96$, RMSE = $0.19$) with stacked model.
 
\end{enumerate}


%----------------------------------------
% 3. Graphical Abstract (on a separate page)
%----------------------------------------
\clearpage
% \section*{Graphical Abstract}
 \begin{figure}
 \centering
\includegraphics[width=\textwidth]{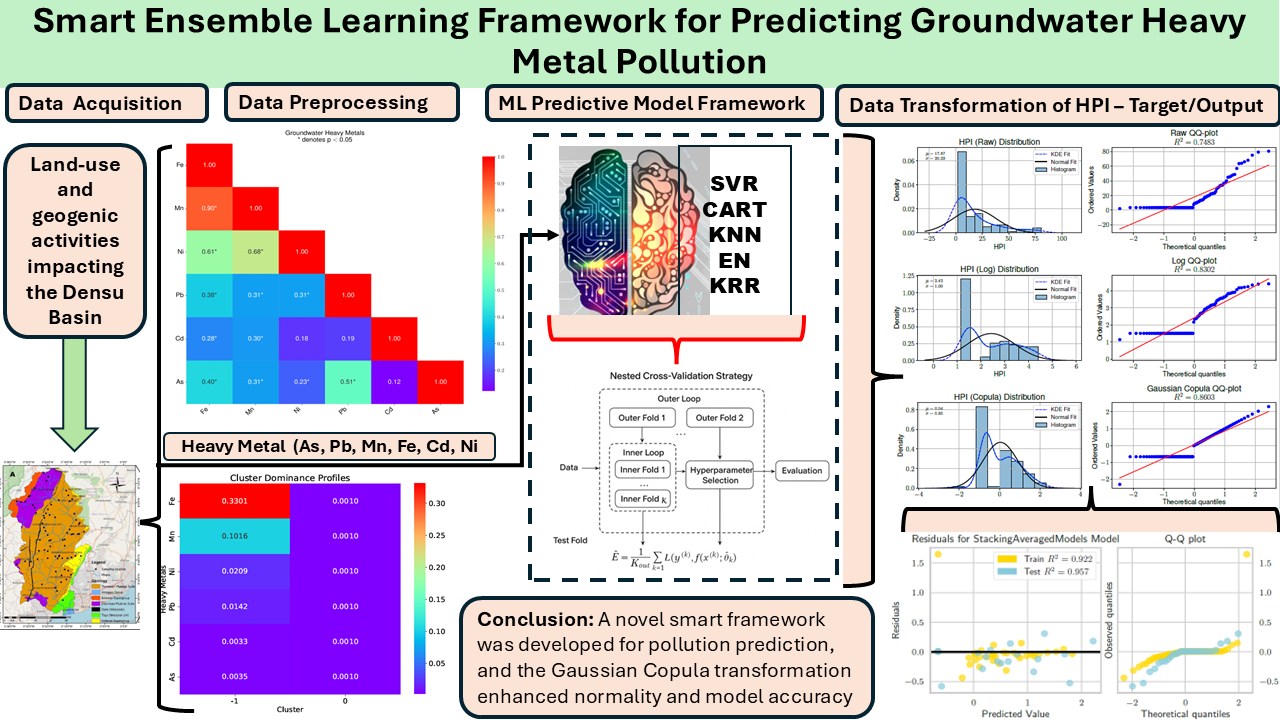}
 \label{Graphical abstract 7.png}
\end{figure}

\textbf{Graphical abstract descriptions:} 
The graphical abstract illustrates the workflow for predicting groundwater heavy metal pollution in the Densu Basin using a nested cross-validated stacked ensemble learning framework. 
It begins with the acquisition and preprocessing of data on six key metals (As, Pb, Mn, Fe, Cd, Ni), highlighting the land-use and geogenic activities that influence contamination. 
Correlation and cluster dominance analyses reveal Fe and Mn as the major contributors to the Heavy Metal Pollution Index (HPI). 
The predictive framework integrates multiple machine learning models (SVR, CART, KNN, Elastic Net, Kernel Ridge) combined through stacking and nested cross-validation to ensure unbiased performance estimation. Comparative transformation of the HPI (raw, log, and Gaussian copula) demonstrates that the Gaussian copula approach improves normality and predictive accuracy. The concluding section summarizes that this framework effectively enhances model reliability and interpretability for groundwater quality prediction, providing a structured, data-driven approach for assessing heavy metal contamination in hydrogeochemical systems.

\maketitle
\section{Introduction} \label{sec:introduction}

Groundwater is a critical resource for drinking water security worldwide,
 particularly in regions like Ghana where surface water availability is often scarce or seasonal.
 Its role is especially vital in rural and peri-urban communities, 
  where it frequently constitutes the primary source for domestic and agricultural supply. 
  However, groundwater quality is increasingly threatened by contamination from both natural geogenic sources
   and anthropogenic activities, including mining, agriculture, and industrial effluent discharge
  \citep{li2018assessment}. 
Among the most persistent and hazardous contaminants are heavy metals, such as lead (Pb), 
    nickel (Ni), cadmium (Cd), iron (Fe), manganese (Mn), and arsenic (As). Due to their toxicity,
     persistence, and capacity for bioaccumulation, these elements pose significant public health risks even 
     at trace concentrations.
 Consequently, robust and efficient frameworks for assessing contamination levels are indispensable 
for proactive water resource management and public health protection.

The Heavy Metal Pollution Index (HPI) is a well-established composite indicator designed to address this need.
It integrates the concentrations of multiple heavy metals into a single, interpretable numerical value, providing a standardised benchmark for evaluating overall water quality relative to permissible limits \citep{eid2024comprehensive,eldaw2020novel,mohan1996estimation}.
As a deterministic formula, the HPI offers a transparent and reproducible metric, making it a valuable tool for environmental scientists and policymakers. 
Its calculation serves as the recognised \enquote{ground truth} for quantifying pollution levels.

However, the practical utility of the HPI for large-scale monitoring programmes is constrained not by its mathematical complexity but by the significant logistical and analytical burdens associated with its data requirements. Its accurate computation is contingent upon the comprehensive laboratory analysis of all requisite heavy metal parameters for each water sample, a process that is financially costly, time-intensive, and requires sophisticated instrumentation and expertise \citep{eldaw2020novel}.
In resource-constrained settings like Ghana, this often results in two pervasive challenges: 

\begin{enumerate}[label=\roman*)]

\item incomplete datasets, where full suites of heavy metal measurements are unavailable for all sampled locations due to budgetary or analytical constraints, and 
\item spatially sparse monitoring networks, where the high cost of sampling leaves vast geographical areas unassessed, hindering a comprehensive basin-wide understanding of pollution risk.
\end{enumerate}

This data scarcity creates a critical operational gap. While geostatistical interpolation techniques available in software like ArcGIS\footnote{It is a proprietary Geographic Information System software suite developed by Esri (Environmental Systems Research Institute).} are commonly used to create spatial surfaces from point data, they are a suboptimal solution for a multivariate index like the HPI. Applying kriging or inverse distance weighting to interpolate each metal concentration individually before calculating the HPI introduces a compounding of errors from each individual model, while simultaneously failing to capture the complex, non-linear interdependencies between the metal parameters that influence the final index value \citep{afrifa2021estimation}. 
Therefore, a more sophisticated approach is needed to translate limited point data into reliable, basin-wide insights.

 Machine learning (ML) offers a powerful paradigm to address this challenge. Rather than simply replicating the HPI calculation, a robustly trained ML model can learn the underlying functional relationship between the input metal concentrations and the resultant HPI value. 
 This capability provides two distinct advantages over conventional methods: firstly, the model can impute accurate HPI values for samples with missing metal parameters, effectively overcoming data gaps. Secondly, and more significantly, it can predict HPI at unmonitored locations by learning from the spatial and multivariate structure of the existing data, enabling the creation of continuous pollution risk maps from sparse point measurements.
 This predictive capacity facilitates a proactive approach to identifying pollution hotspots, optimising future sampling campaigns, and guiding mitigation efforts, applications that are prohibitively difficult with manual HPI computation alone.

The urgency of developing such tools is underscored by evidence of heavy metal contamination in Ghana's water resources. Studies in basins like the Densu have reported concentrations of Pb, Ni, and As exceeding World Health Organization (WHO) guidelines, linked to geological sources and exacerbated by anthropogenic pressures such as artisanal mining and intensive agriculture \citep{osei2023assessments,amoako2011physico,tay2008groundwater}.
The health implications of chronic exposure to metals like arsenic and lead are severe, including increased risks of cancers, neurological impairment, and cognitive deficits in children \citep{speer2023arsenic,world2022guidelines,zietz2001lead}.
This reality elevates the need for efficient assessment tools from an academic exercise to a public health priority.

In this context, the present study proposes a novel nested cross-validated stacked ensemble regression framework to predict the HPI directly from groundwater heavy metal concentrations. We move beyond the question of computational efficiency to address the more pertinent challenges of data scarcity and spatial prediction. To enhance model generalisability and explore the impact of data representation, we assess the efficacy of Gaussian Copula transformation against raw and log-transformed inputs for preserving multivariate dependencies. The performance of this predictive framework is rigorously benchmarked against the deterministically calculated HPI. 
Note that our objective is not to replace the HPI but to develop a complementary, robust, and scalable predictive tool that augments its utility, enabling more comprehensive and timely groundwater quality assessment for the Densu River Basin and other data-scarce regions globally.

The remainder of this paper is organised as follows: Section~\ref{sec:study_area} describes the study area and sampling context. Section~\ref{sec:methods} outlines the data acquisition, preprocessing, HPI computation, response transformations, and the nested cross-validated ensemble modelling framework. Section~\ref{sec:geospatial} details the geospatial interpolation procedures used to generate grid-level predictors. Section~\ref{sec:metrics} defines the performance metrics and diagnostic procedures. Section~\ref{sec:results} presents and discusses the results, including prediction-error analysis, residual diagnostics, clustering-based dominance assessment, and spatial HPI predictions under alternative transformations. Finally, Section~\ref{sec:Conc} summarises the findings, discusses limitations and management implications, and outlines future research directions.

% \section{Study Area} 
% \begin{enumerate}[label=\roman*.]
%     \item quantify the impact of these climate indices on local weather variability, 
%     \item assess the temporal and spatial patterns of these teleconnections, and 
%     \item evaluate the implications of these findings for climate adaptation strategies in northern Ghana.
% \end{enumerate}
% By shedding light on the linkages between remote climate drivers and local climate conditions, this study will contribute to a more comprehensive understanding of regional climate dynamics, aiding policymakers and stakeholders in crafting effective climate adaptation and resilience strategies for northern Ghana.

\section{Study Area} \label{sec:study_area}

\begin{figure}
\includegraphics[width=\textwidth]{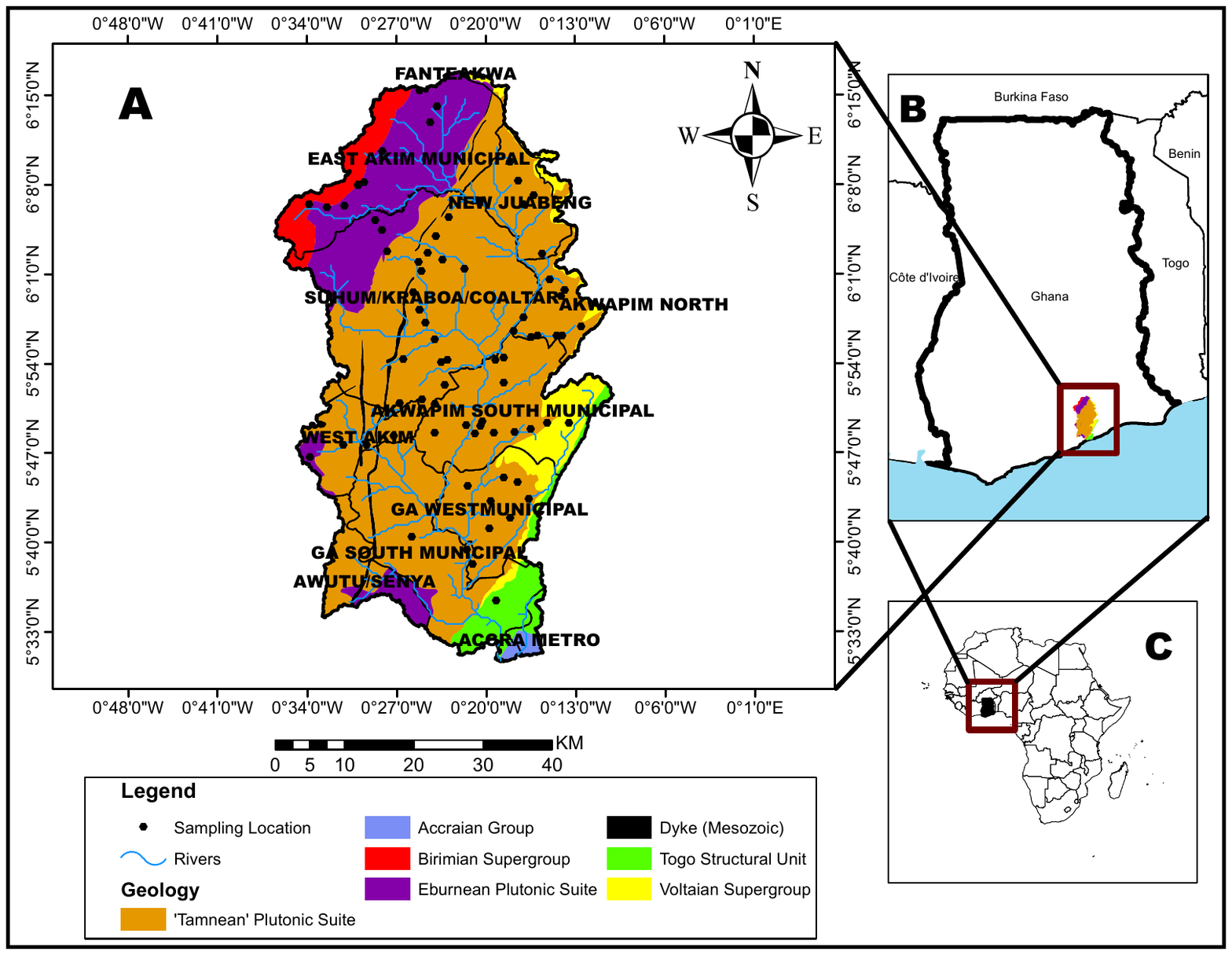}
\caption{Map of the study area showing sampling locations, major rivers, and underlying geology within parts of southern Ghana. The geological units include the Birimian Supergroup, Tarkwaian Group, Togo Structural Unit, Voltaian Supergroup, and various plutonic suites, providing context for groundwater hydrochemistry and heavy metal distribution.}
\label{fig:studyAreamap}
\end{figure}

The Densu River Basin is located in southern Ghana between latitudes $5^\circ 30'\,\mathrm{N}$ and $6^\circ 20'\,\mathrm{N}$, and longitudes $0^\circ 10'\,\mathrm{W}$ and $0^\circ 35'\,\mathrm{W}$ (refer to Fig.~\ref{fig:studyAreamap}).
It spans three administrative regions: the Eastern Region ($72\%$), the Greater Accra Region ($23\%)$, and the Central Region ($5\%$) \citep{nyamekye2016using}. 
The basin is hydrologically and socio-economically significant, supplying potable water to major urban centres, including Accra, while sustaining extensive agricultural activity. 
Climatically, it exhibits a bimodal rainfall regime, with a wet semi-equatorial climate dominating the upper basin and a dry equatorial climate prevailing in the southeast.

Agriculture is the dominant land use, encompassing both food crops such as maize, cassava, plantain, yam, banana, and cocoyam, and cash crops including pineapple, oil palm, cocoa, papaya, mango, and citrus \citep{amoako2011physico}.
Additional land uses include housing development, sand winning, animal rearing, and salt mining. These activities have contributed to significant vegetation loss and have exacerbated hydrological and geomorphological issues such as flooding, soil erosion, siltation, and increased evaporation.

Geologically, the basin comprises diverse lithological units, including the Tamnean Plutonic Suite, Tarkwaian and Birimian Supergroups, Accraian Group, Eburnean Plutonic Suite, Togo Structural Unit, and Voltaian Supergroup, along with intrusive dykes \citep{amoako2011physico,akurugu2022groundwater}. These formations influence groundwater geochemistry and aquifer characteristics, and in some cases facilitate the mobilisation of trace metals. Seasonal rainfall patterns and land use changes further increase the potential for contaminant transport into groundwater systems.

Anthropogenic pressures, including artisanal mining, agrochemical application, and urban effluent discharge, place additional stress on the basin’s aquifers \citep{tay2008groundwater,amoako2011physico}.
The combination of geological susceptibility and human-induced stressors makes the Densu River Basin an ideal case study for developing and validating advanced predictive models for heavy metal pollution assessment.

\section{Materials and Methods} \label{sec:methods}
\subsection{Data acquisition} \label{sec:collection}

This study utilised a hydrochemical dataset comprising concentrations of six heavy metals: Pb, Ni, Cd, Fe, Mn, and As, quantified in groundwater samples.
Sampling was conducted over a targeted campaign in January 2020 across the Densu River Basin. A total of 96 sampling points were selected through a stratified random approach to ensure representative coverage of the basin's major aquifer systems and land use types. The sampled sources encompassed both mechanised boreholes and hand-dug wells, which constitute the primary drinking water infrastructure for local communities, thereby ensuring the practical relevance of the data.

A strict sampling protocol, consistent with international standards for trace metal analysis (e.g., Standard Methods for the Examination of Water and Wastewater\footnote{\url{https://www.standardmethods.org/}}), was adhered to in order to preserve sample integrity. Groundwater was collected in pre-cleaned 500 mL low-density polyethene bottles, which were triple-rinsed with the sample water prior to collection. Immediately following collection, each sample was acidified with $1$ mL of ultrapure nitric acid (HNO\textsubscript{3}, Trace Metal Grade\footnote{\url{https://documents.thermofisher.com/TFS-Assets/GC/brochures/fisher-chemical-high-purity-acid.pdf}}) to a pH of less than $2$, thereby preventing cation adsorption and precipitation. Samples were immediately placed in insulated coolers maintained at below \SI{4}{\degreeCelsius} to inhibit microbial activity and minimise volatilisation. Under this cold chain, samples were transported to the central analytical laboratory at the National Nuclear Research Institute of the Ghana Atomic Energy Commission \footnote{\url{https://gaec.gov.gh/}}.

Before instrumental analysis, samples underwent a digestion process to decompose organic-metallic complexes and release total metal concentrations. 
A $100$ mL aliquot of each sample was digested with $5$ mL of concentrated HNO\textsubscript{3} on a hot plate at \SI{95}{\degreeCelsius} until the volume was reduced to approximately $50$ mL, ensuring complete dissolution of the target analytes. The digested samples were then cooled, filtered, and made up to the original volume with deionised water.

The concentrations of As, Cd, Pb, Ni, Fe, and Mn were determined using flame atomic absorption spectrophotometry, following established standard methods. Quality assurance and quality control protocols were rigorously implemented. This included the analysis of procedural blanks, duplicate samples, and certified reference materials (CRMs) with each batch of $20$ samples to monitor for contamination, precision, and accuracy, respectively. The recovery rates for the CRMs ranged between $90\%$ and $105\%$, confirming the reliability of the analytical data. This stringent procedure minimised potential contamination and ensured the generation of robust concentration data, which served as the fundamental input for both the deterministic computation of the HPI and the subsequent training of the MLs.

\subsection{Data preprocessing and exploratory analysis}
\label{subsec:prep}

\subsubsection{Summary statistics}

Table~\ref{tab1} provides descriptive statistics of the heavy metal concentrations in groundwater samples collected from the study area. All metals exhibit a marked right-skewed distribution, with means consistently higher than medians, and standard deviations that are often greater than the means themselves.
For example, Fe recorded a mean of $0.169$ mg L$^{-1}$ but a median equal to the detection limit ($0.001$ mg L$^{-1}$), accompanied by a maximum concentration of $2.56$ mg L$^{-1}$. A similar though less pronounced trend was observed for Mn, which had a mean of $0.052$ mg L$^{-1}$ and a maximum of $0.25$ mg L$^{-1}$. 
Other metals such as Ni, Pb, Cd, and As showed low means but also displayed extreme maximum values relative to their medians. These statistics indicate that most samples were at or near the analytical detection limits, but that a small number of samples exhibited elevated concentrations well above typical background levels.

The statistical distribution observed in this dataset is typical of groundwater systems influenced by both geogenic and anthropogenic processes.
Fe and Mn are widely reported to co-occur in reducing aquifers where the reductive dissolution of Fe/Mn oxyhydroxides mobilises these elements simultaneously \citep{Saeed2023,Zhai2022}.
The elevated maximum values for Pb and As, despite very low median levels, may suggest localised anthropogenic sources such as agricultural inputs and waste disposal, in addition to possible contributions from quarrying and sand-winning activities in the basin.
The highly skewed distribution, together with the accumulation of values at the analytical detection limit, indicates that many concentrations fall below the measurable threshold of the instruments. This censoring effect complicates the direct application of parametric statistical methods, making robust and non-parametric approaches, such as rank-based statistics, more suitable for examining the relationships among the measured elements \citep{Akoglu2018}.

\begin{table}
\centering
\caption{Descriptive statistics of heavy metal concentrations in groundwater samples from the study area.}\label{tab1}
\begin{threeparttable}
\begin{tabular}{@{}l *{6}{S[table-format=1.3]} @{}}
\toprule
Statistical & {Fe} & {Mn} & {Ni} & {Pb} & {Cd} & {As}  \\
Parameters & & & & & & \\
\midrule
Count & 96.00 & 96. & 96 & 96 & 96 & 96 \\
Mean & 0.169 & 0.052 & 0.011 & 0.007 & 0.002 & 0.002 \\
Std & 0.309 & 0.065 & 0.018 & 0.016 & 0.004 & 0.003 \\
Min & 0.001 & 0.001 & 0.001 & 0.001 & 0.001 & 0.001 \\
25\% & 0.001 & 0.001 & 0.001 & 0.001 & 0.001 & 0.001 \\
50\% & 0.001 & 0.001 & 0.001 & 0.001 & 0.001 & 0.001\\
75\% & 0.25 & 0.089 & 0.014 & 0.001 & 0.001 & 0.001\\
Max & 2.56 & 0.25 & 0.088 & 0.072 & 0.032 & 0.020\\
\bottomrule
\end{tabular}
\begin{tablenotes}
\small
\item Note: All metal concentrations are in mg/L.
\end{tablenotes}
\end{threeparttable}
\end{table}

\subsubsection{Treatment of values at the reporting limit}
A substantial proportion of observations are reported at 0.001 mg L$^{-1}$, corresponding to the laboratory reporting limit. In this study, such values were retained as recorded and not treated as missing or left-censored observations. This reflects the analytical protocol, where values at the reporting limit represent quantifiable concentrations rather than non-detects.

No substitution (e.g., LOD/2 or LOD/$\sqrt{2}$) or model-based imputation was applied, as these approaches can introduce bias in skewed environmental data and distort rank-based dependence structures. Retaining reported values preserves consistency with quality-controlled laboratory outputs and maintains the empirical ordering required for subsequent non-parametric procedures.

For transformations, numerical stability was ensured by applying $y^{*}=\log(1+y)$, so values at or near the reporting limit remain well-defined. For the Gaussian copula transformation, ranks were computed directly from the observed data, with ties handled using average ranks, thereby preserving monotonic relationships without imposing parametric assumptions. The HPI was computed deterministically from these reported concentrations, ensuring full traceability between analytical measurements and derived index values.

\subsubsection{Correlation analysis}\label{sec:corr}

Given the skewed distributions and presence of censoring in the dataset, Spearman’s rank correlation coefficient was adopted to assess monotonic associations between pairs of heavy metals. Unlike Pearson’s correlation, Spearman’s method relies on the ranks of the data rather than their raw values, reducing sensitivity to extreme outliers and distributional departures from normality. Let $X = \{x_1, x_2, \ldots, x_n\}$ and $Y = \{y_1, y_2, \ldots, y_n\}$ denote two random variables with $n$ paired observations. The values are converted to ranks $R(x_i)$ and $R(y_i)$, with ties assigned the average rank. Spearman’s correlation coefficient is then given as:

\begin{equation}
\rho_s = 1 - \frac{6 \sum_{i=1}^n d_i^2}{n(n^2-1)},
\end{equation}

\noindent where $d_i = R(x_i) - R(y_i)$. 
When ties are present, $\rho_s$ is equivalently defined as the Pearson correlation of the rank-transformed data:

\begin{equation}
\rho_s = \frac{\sum_{i=1}^n (R(x_i) - \overline{R_x})(R(y_i) - \overline{R_y})}{\sqrt{\sum_{i=1}^n (R(x_i) - \overline{R_x})^2 \sum_{i=1}^n (R(y_i) - \overline{R_y})^2}}.
\end{equation}

The null hypothesis tested is that no monotonic association exists between the two variables:

\begin{equation}
H_0: \rho_s = 0 \quad \text{versus} \quad H_a: \rho_s \neq 0.
\end{equation}

\noindent An approximate test statistic can be constructed as:

\begin{equation}
t = \rho_s \sqrt{\frac{n-2}{1-\rho_s^2}},
\end{equation}

\noindent which under $H_0$ follows approximately a Student’s $t$ distribution with $(n-2)$ degrees of freedom. However, given the strong skewness and high frequency of ties in the present dataset, permutation-based inference was also considered more robust. In this case, the empirical distribution of $\rho_s$ is obtained by repeatedly permuting the $Y$ values while keeping $X$ fixed, and recalculating $\rho_s$ for each permutation. The permutation p-value is then computed as:

\begin{equation}
p_{\text{perm}} = \frac{1 + \#\{b : \lvert \rho_s^{(b)} \rvert \geq \lvert \rho_s^{\text{obs}} \rvert\}}{1+B},
\end{equation}

\noindent where $\rho_s^{\text{obs}}$ is the observed correlation, $\rho_s^{(b)}$ is the correlation under the $b$-th permutation, and $B$ is the total number of permutations \citep{Hutson2023,Yu2022}. The symbol ($\#$) denotes the cardinality of the set, thus, the number of permutations for which the condition inside the braces holds. 
Note that the permutation test essentially compares the observed statistic to the empirical null distribution obtained by permuting the data: the p-value is the proportion of permuted correlations that are at least as extreme as the observed one.

The correlation analysis of the groundwater dataset revealed several important relationships.
Fe and Mn exhibited the strongest correlation ($\rho_s = 0.90$, $p < 0.05$), consistent with their shared mobilisation under reducing conditions typical of aquifers in the study area. 
Ni showed moderate to strong correlations with both Fe and Mn ($\rho_s = 0.61$ and $\rho_s = 0.68$, respectively), suggesting potential co-mobilisation from geogenic mineral phases or anthropogenic effluents. 
Pb and As also displayed a moderate positive association ($\rho_s = 0.51$, $p < 0.05$), which may be explained by their common release from Fe oxyhydroxides or co-introduction from mining-related activities in the region. Cd, on the other hand, showed only weak and mostly non-significant correlations with different metals, indicating more heterogeneous sources and a spatially variable distribution.

\begin{figure}
\includegraphics[width=\textwidth]{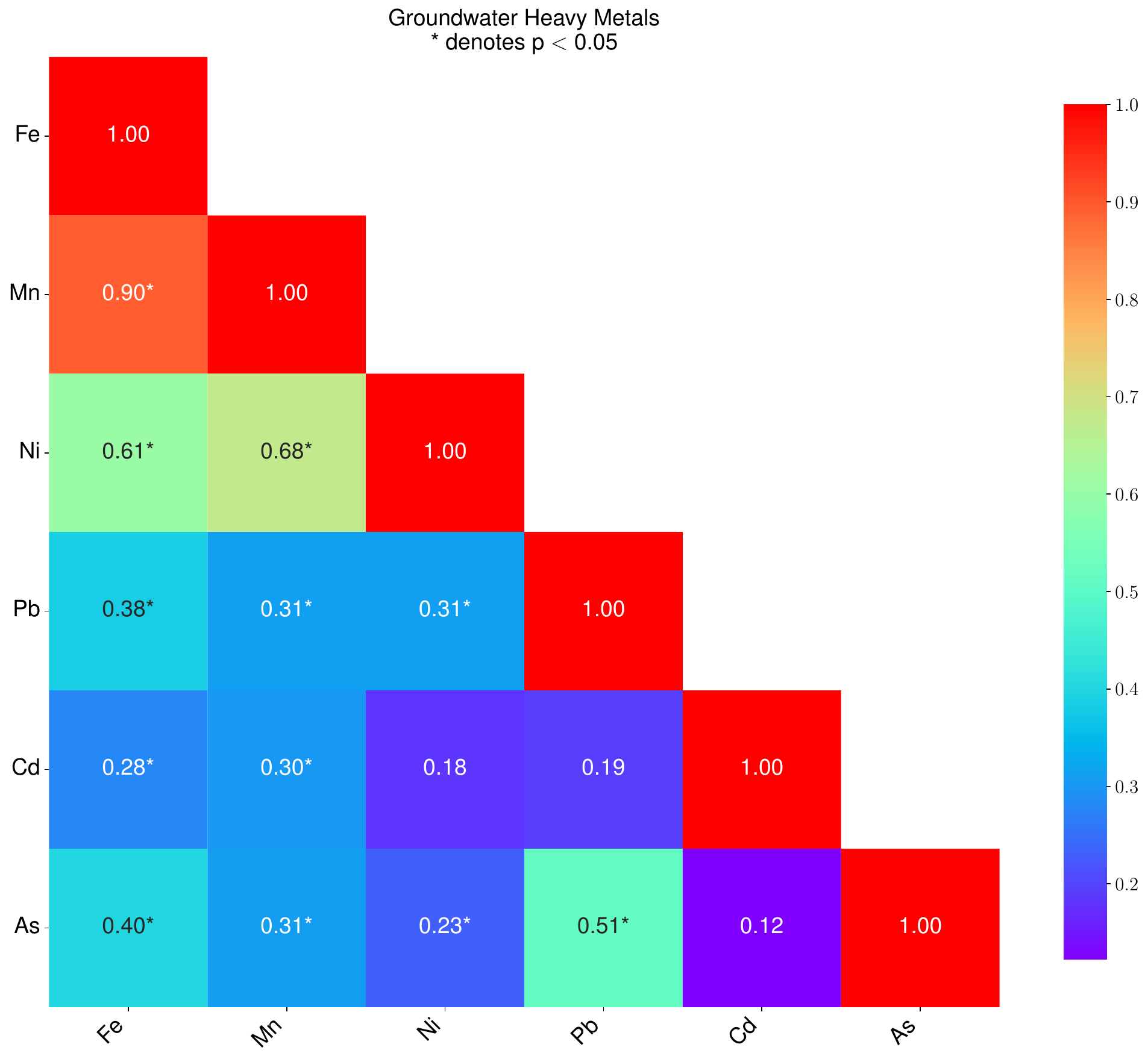}
\caption{Spearman rank correlation matrix of heavy metal concentrations in groundwater samples from the Densu basin. Correlations were computed using rank-transformed data with tie-adjusted Spearman’s $\rho$. Asterisks denote statistically significant correlations at the 5\% level ($p < 0.05$).}
 \label{fig:corr-analysis_raw}
\end{figure}

\subsubsection{Dominance analysis} \label{sec:dscan}

Beyond correlation-based assessments, it is essential to examine the dominance structure of individual metals in order to identify which contaminants exert disproportionate influence within the groundwater system. In this study, a clustering-based dominance analysis was implemented using the density-based spatial clustering of applications with noise (DBSCAN) algorithm. DBSCAN is particularly well-suited to groundwater data, as it can identify clusters of arbitrary shape while distinguishing noise points, which may correspond to localised pollution outliers \citep{kazemi2025new,ester1996density}. To ensure comparability, raw concentration values were standardised using $z$-score transformation prior to clustering \citep{Jain2010}. Following clustering, centroids were transformed back to original concentration units to facilitate interpretation in hydrogeochemical terms.

The procedure can be formalised as follows: let $X \in \mathbb{R}^{n \times p}$ denote the groundwater dataset, where $n$ is the number of samples and $p$ the number of metals analysed. After scaling, the dataset is partitioned into $C$ clusters identified by DBSCAN, yielding centroid vectors $\mu_c \in \mathbb{R}^p$, $c = 1, \ldots, C$. The dominant metal for cluster $c$ is then identified as
\begin{equation}
M_c = \arg\max_{j \in \{1,\ldots,p\}} \mu_{c,j},
\label{eq:dominant_metal}
\end{equation}
where $\mu_{c,j}$ denotes the mean concentration of metal $j$ within cluster $c$. This criterion (in Eq.~\eqref{eq:dominant_metal}) provides an interpretable and systematic method to establish which metal is most characteristic of a given hydrogeochemical domain.
The DSCAN procedure used in this work is outlined in Algorithm~\ref{alg:dscan}.

\begin{algorithm}[H]
\caption{Dominance Analysis of Heavy Metals using DSCAN}\label{alg:dscan}
\begin{algorithmic}[1]
\Require Groundwater dataset $X \in \mathbb{R}^{n \times p}$, DBSCAN parameters $(\varepsilon, \text{min\_samples})$
\Ensure Dominant metals per cluster $\{M_c\}_{c=1}^C$
\State Standardise $X$ using $z$-score transformation
\State Apply DBSCAN with parameters $\varepsilon$, min\_samples
\State Assign cluster labels to all samples
\For{each cluster $c$}
    \State Compute centroid $\mu_c$ in scaled space
    \State Inverse-transform $\mu_c$ to original concentration units
    \State Determine dominant metal $M_c = \arg\max_j \mu_{c,j}$
\EndFor
\State \Return $\{M_c\}_{c=1}^C$ and cluster-level centroid profiles
\end{algorithmic}
\end{algorithm}

%% Cluster plot
\begin{figure}[H]
\centering
\includegraphics[width=\textwidth]{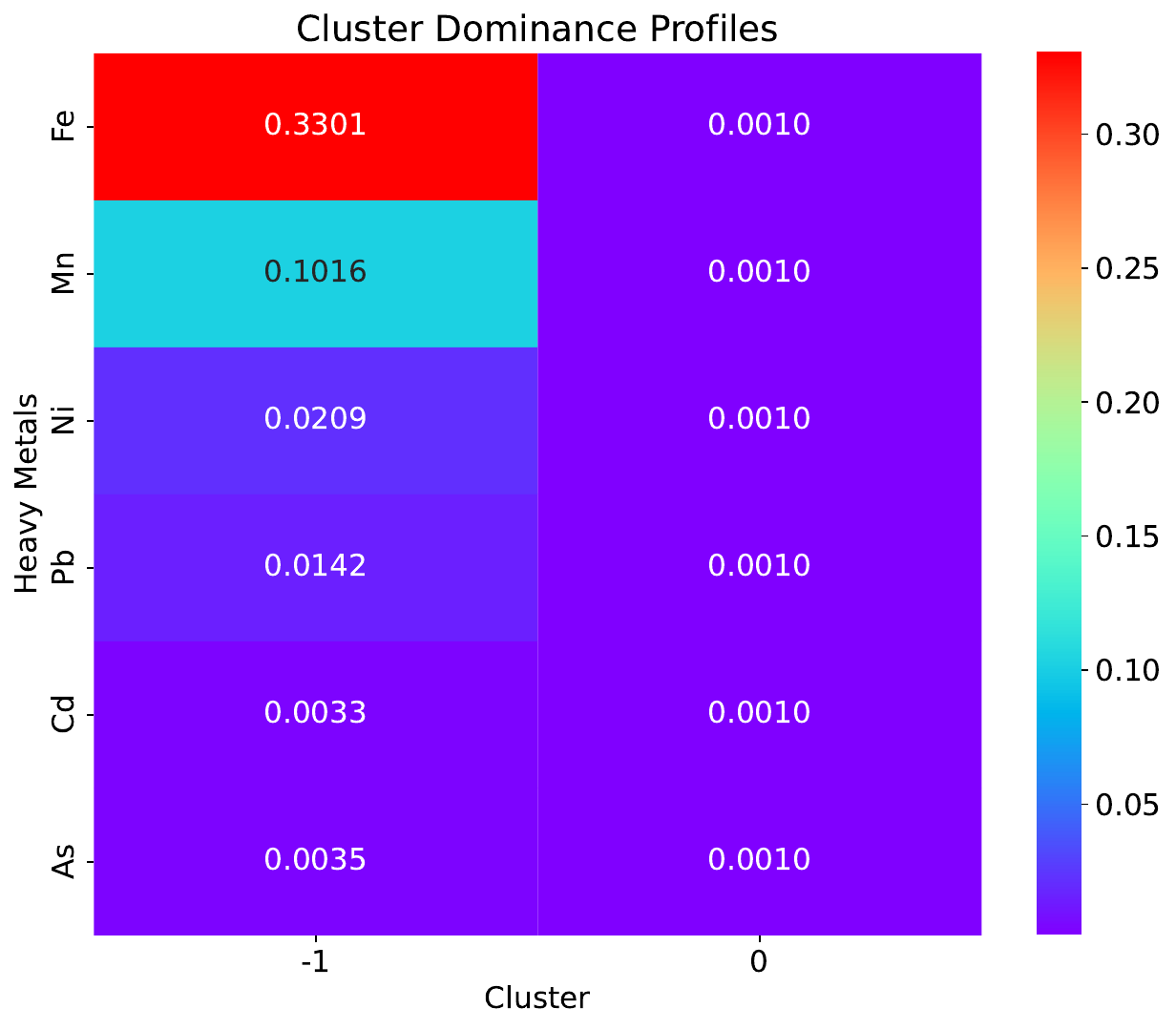}
\caption{The cluster dominance profile in the Densu Basin groundwater, showing centroid values for Fe, Mn, Ni, Pb, Cd, and As across identified clusters. Cluster $-1$ is characterised by elevated concentrations with Fe as the dominant metal, while Cluster $0$ represents background conditions with near-minimal levels across all metals.}
\label{fig:clusterPlot}
\end{figure}

The cluster dominance profile in Fig.~\ref{fig:clusterPlot} highlights two distinct hydrogeochemical regimes within the Densu Basin groundwater.
 Cluster $-1$ exhibits markedly higher centroid values across all metals, with Fe emerging as the most dominant constituent ($0.3301$ mgL\textsuperscript{-1}). This pattern is consistent with the known geochemical behaviour of Fe in tropical aquifers, where redox fluctuations drive the mobilisation of Fe oxyhydroxides and associated trace elements \citep{Smedley2002}. 
 Mn also displays a substantial centroid value ($0.1016$ mgL\textsuperscript{-1}), reinforcing its role as a co-occurring redox-sensitive element. The relatively elevated presence of Ni, Pb, Cd, and As, although at lower magnitudes, suggests that these metals may be mobilised secondarily, either through sorption–desorption processes or co-release from Fe–Mn mineral phases \citep{zuniga2019evaluation,lee2002removal,mclean1992behavior}.

 To ensure that this dominance structure is not an artefact of clustering configuration, the DBSCAN algorithm was implemented with $\varepsilon = 0.5$ and $\text{min\_samples} = 10$ on z-score standardised metal concentrations. Under standardisation, $\varepsilon$ represents a moderate density threshold in the transformed feature space, while $\text{min\_samples}$ (approximately 10\% of the dataset) ensures that identified clusters reflect statistically meaningful groupings rather than noise. Sensitivity checks over a reasonable range of $\varepsilon$ and $\text{min\_samples}$ values showed that, although minor variations in cluster membership may occur, the dominance ordering of metals, particularly the consistent identification of Fe and Mn as primary contributors, remains unchanged. This indicates that the hydrogeochemical interpretation is robust to DBSCAN parameter selection.

\subsection{Computation of the heavy metal pollution index}

The HPI is a composite metric that provides an integrated assessment of water quality with respect to concurrent heavy metal contamination. Its core utility lies in its ability to aggregate complex multi-parameter data into a single, interpretable value, facilitating straightforward comparison and prioritisation of polluted sites \citep{Singh2019,mohan1996estimation}.
In this study, the HPI was computed deterministically for each groundwater sample according to the established weighted arithmetic mean method to serve as the ground-truth target variable for subsequent machine learning modelling.

The calculation procedure involves three primary steps for each of the $n$ heavy metal parameters:

\begin{enumerate}[label=\roman*)]

\item Calculation of the sub-index ($Q_i$) for each metal, which represents its relative contribution to pollution:

\begin{equation} \label{eq:subindex}
Q_i = \frac{M_i - I_i}{S_i - I_i} \times 100
\end{equation}

\noindent where $M_i$ is the measured concentration of the $i$-th metal, $S_i$ is its permissible limit as defined by the World Health Organisation \citep{world2022guidelines} guidelines for drinking water,  and $I_i$ is its ideal value. 
In line with common practice for harmful contaminants, the ideal value $I_i$ was set at zero for all metals, reflecting the optimal scenario of their absence in drinking water \citep{eldaw2020novel}.

\item Assignment of the unit weight ($W_i$) to each parameter, which is inversely proportional to its recommended standard:

\begin{equation} \label{eq:unit_weight}
W_i = \frac{k}{S_i}
\end{equation}

\noindent  where $k$ is a proportionality constant. A value of $k=1$ was used herein, ensuring that a metal with a lower permissible limit (e.g., Cd or As) exerts a greater influence on the final index than a metal with a higher limit (e.g., Fe or Mn), reflecting its higher relative toxicity \citep{Rezaei2019}.

\item Aggregation into the final HPI value via the weighted mean of all sub-indices:

\begin{equation} \label{eq:agg}
\text{HPI} = \frac{\sum_{i=1}^n (W_i \cdot Q_i)}{\sum_{i=1}^n W_i}
\end{equation}

\end{enumerate}

\noindent The resulting HPI values were interpreted using a well-established classification scheme wherein water quality is categorised as excellent (HPI $<$ 15), good to intermediate (15 $\le$ HPI $\le$ 30), poor to unsuitable (31 $\le$ HPI $\le$ 75), very poor (76 $\le$ HPI $\le$ 100), or unsuitable for drinking (HPI $>$ 100), in accordance with standard practices for heavy metal pollution assessment \citep{Rezaei2019,mamat2016ecological}. 
This deterministic HPI, derived directly from the measured hydrochemical data, served as the foundational benchmark against which the predictive performance of all ML models was rigorously evaluated.

\subsection{Predictive modelling framework}

\subsubsection{Data partitioning and feature--target separation}

The predictive modelling framework was constructed by distinguishing between predictor variables ($X$) and the target variable ($y$). In this study, the predictor matrix $X$ comprised the concentrations of six heavy metals (Fe, Mn, Ni, Pb, Cd, and As) measured in groundwater samples from the Densu basin, while the response variable $y$ was the computed HPI. 
The partitioning procedure followed standard practice in machine learning to enable model training, validation, and assessment of generalisation capacity \citep{Kuhn2013}. Specifically, the dataset was split into training and testing sets using a $70{:}30$ ratio through random sampling, ensuring that all predictors and the target were proportionally represented in both partitions. 

To improve model stability and comparability between variables measured at different scales, $X$ was standardised using the \texttt{StandardScaler} transformation. This operation centres each predictor around zero mean and scales it to unit variance, thereby mitigating the risk of predictors with larger absolute magnitudes disproportionately influencing the learning process \citep{James2021}. Formally, for predictor $x_{ij}$ representing the $j$-th variable in the $i$-th observation, the transformation is given as:

\begin{equation}
x_{ij}^{\ast} = \frac{x_{ij} - \mu_j}{\sigma_j},
\end{equation}

\noindent where $\mu_j$ and $\sigma_j$ are the sample mean and standard deviation of the $j$-th predictor in the training set. The scaled predictors ($X^{\ast}$) and the partitioned response ($y$) then served as inputs to subsequent baseline and ensemble learning algorithms described in Section~\ref{sec:baselines}.

\subsubsection{Data transformation techniques}

\begin{figure}[H]
\centering
\includegraphics[width=0.95\textwidth]{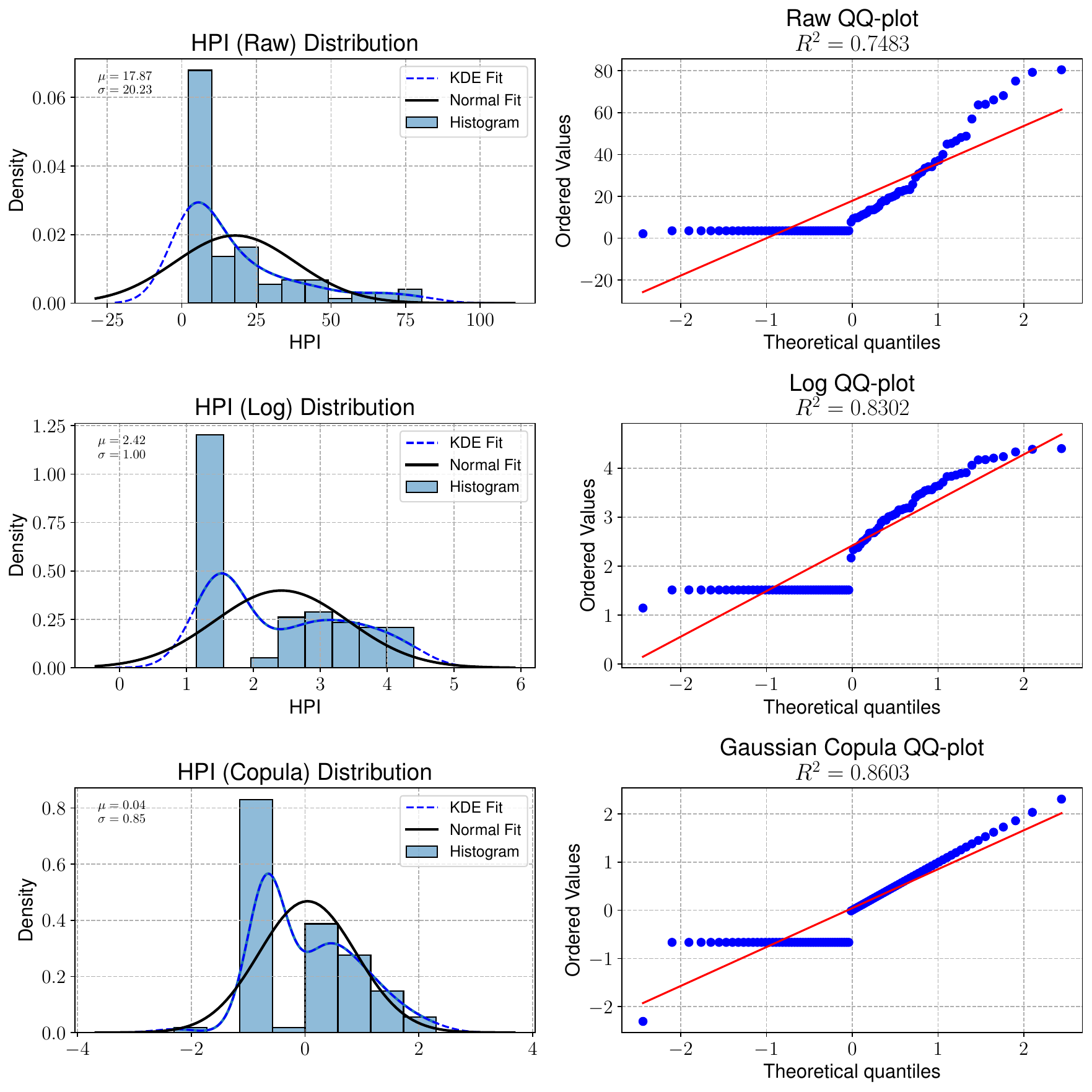}
\caption{Distributional diagnostics of the HPI under different transformations. Left column: empirical histograms with fitted kernel density estimates (KDE) and normal densities; right column: quantile--quantile (QQ) plots with fitted lines and $R^2$ statistics assessing goodness-of-fit to the normal distribution. The Gaussian copula transformation yields the closest approximation to normality while preserving dependence structure.}
\label{fig:HPI_normality}
\end{figure}

The HPI was subjected to three distinct treatments prior to modelling: raw scale, logarithmic transformation, and Gaussian copula transformation. This choice was motivated by the empirical distributional properties of the HPI, which exhibited strong positive skewness and significant departure from normality (Fig.~\ref{fig:HPI_normality}). Such features are typical of environmental quality indices derived from skewed pollutant concentration data. 
The three transformations were evaluated to explore the extent to which normalisation of marginal distributions influences predictive performance.

In the first case, the raw HPI values were used directly as the target variable. This approach preserves the original interpretability of the index but suffers from distributional irregularities such as skewness, which can reduce predictive accuracy for models sensitive to distributional assumptions \citep{James2021,Sohil2021}.

The second approach applied a natural logarithmic transformation, $y^{\ast} = \log(1+y)$, which is widely used to stabilise variance, compress extreme values, and approximate normality in positively skewed environmental data.
The addition of one ensures positivity and prevents undefined values when $y=0$. The log transformation often improves model fit when the response distribution is heavy-tailed or contains extreme outliers.

The third approach employed a Gaussian copula transform, a non-parametric method that maps a continuous marginal distribution to a standard normal distribution while preserving dependence structure \citep{Krupskii2018}. The procedure involves three steps: 
\begin{enumerate}[label=(\roman*)]
    \item Rank transformation,
    \item Mapping ranks to the uniform distribution,
    \item Applying the inverse Gaussian cumulative distribution function (CDF).
\end{enumerate} 
For an observed response vector $\{y_1, y_2, \ldots, y_n\}$, let $R(y_i)$ denote the rank of $y_i$. The uniform scores are defined as
\begin{equation}
u_i = \frac{R(y_i)}{n+1}, \qquad i=1,\ldots,n,
\end{equation}
\noindent where $n$ is the sample size. These are then mapped to the standard normal scale via the probit transformation
\begin{equation}
z_i = \Phi^{-1}(u_i),
\end{equation}
\noindent where $\Phi^{-1}(\cdot)$ is the quantile function of the standard normal distribution. The transformed variable $\{z_i\}$ is approximately Gaussian (Fig.~\ref{fig:HPI_normality}) while preserving the rank-based dependence structure of $y$. This provides a tractable representation for modelling while retaining the robustness of rank information.

While copula methods are typically used to construct joint distributions by transforming all variables to uniform margins, here the transformation is applied only to the response. The objective is to model the conditional distribution of HPI given metal concentrations, rather than the full joint distribution. Transforming only the response stabilises variance and improves distributional properties without altering the predictor space. This choice is motivated by three considerations:    
\begin{enumerate}[label=(\roman*)]
    \item metal concentrations retain physical interpretability in their original units;
    \item the employed algorithms, particularly tree-based and instance-based methods, are largely insensitive to predictor scaling and distribution; and
    \item the primary statistical issue lies in the skewed response rather than the predictors.
\end{enumerate} 
This approach is analogous to a Box--Cox transformation of the target, but extends naturally to arbitrary distributions through rank-based mapping followed by the inverse Gaussian CDF.

The comparative plots in Fig.~\ref{fig:HPI_normality} confirm the efficacy of the Gaussian copula approach in approximating normality, as reflected by higher $R^2$ values in the QQ-plot diagnostics. This transformation thus addresses the dual challenge of non-normality and the need to preserve monotonic relationships between the HPI and its predictors. By contrast, the log transformation reduced skewness but was insufficient to achieve Gaussian margins, while the raw distribution exhibited strong departures from normality.

\subsubsection{Baseline and ensemble models} \label{sec:baselines}

The modelling framework began with a set of baseline regressors representing diverse algorithmic families: linear, kernel-based, tree-based, and instance-based methods. The intention was to capture complementary strengths in linear approximation, non-linear mapping, hierarchical partitioning, and local neighbourhood search \citep{Hastie2009, James2021}. The baseline models included support vector regression (SVR), decision tree regression (CART), $k$-nearest neighbours (kNN), Elastic Net regression, and kernel ridge regression (KRR).

SVR extends the principles of support vector machines to regression tasks. Given training data $\{(x_i, y_i)\}_{i=1}^n$ with $x_i \in \mathbb{R}^p$ and $y_i \in \mathbb{R}$, SVR seeks a function $f(x) = \langle w, \phi(x)\rangle + b$ that approximates $y_i$ with an $\varepsilon$-insensitive loss. The optimisation problem is:

\begin{equation}
\min_{w,b,\xi_i,\xi_i^\ast} \frac{1}{2}\lVert w \rVert^2 + C \sum_{i=1}^{n} (\xi_i + \xi_i^\ast), 
\label{eq:svr}
\end{equation}

\noindent subject to the constraints
\[
y_i - \langle w,\phi(x_i)\rangle - b \leq \varepsilon + \xi_i, \quad 
\langle w,\phi(x_i)\rangle + b - y_i \leq \varepsilon + \xi_i^\ast, \quad 
\xi_i, \xi_i^\ast \geq 0.
\]

\noindent Here, $\phi(\cdot)$ is a feature map, $C>0$ controls the trade-off between margin flatness and training error, and $\varepsilon$ defines the tolerance zone \citep{Smola2004, drucker1996support}. Thus, Eq.~\eqref{eq:svr} formalises SVR’s ability to achieve sparse solutions that generalise well even in high-dimensional predictor spaces.

Decision tree regression (CART) recursively partitions the input space into regions $R_m$ and fits a constant prediction $c_m$ in each region. The predicted value is given by:

\begin{equation}
f(x) = \sum_{m=1}^{M} c_m \mathbb{I}(x \in R_m),
\label{eq:cart}
\end{equation}

\noindent where $\mathbb{I}(\cdot)$ is the indicator function. As Eq.~\eqref{eq:cart} indicates, the model prediction depends only on the partition $R_m$ into which $x$ falls. Splits are chosen to minimise the sum of squared residuals within nodes, yielding interpretable yet potentially high-variance models \citep{brieman1984classification}.

The $k$-nearest neighbours method predicts a new instance by averaging the outcomes of its $k$ closest neighbours under a chosen distance metric. Formally, the predicted value for $x$ is:

\begin{equation}
\hat{y}(x) = \frac{1}{k} \sum_{i \in N_k(x)} y_i,
\label{eq:knn}
\end{equation}

\noindent where $N_k(x)$ denotes the index set of the $k$ nearest neighbours of $x$ in the training data \citep{Altman1992}. Eq.~\eqref{eq:knn} highlights the local nature of the prediction, though the method can be sensitive to dimensionality effects.

Elastic Net regression is a penalised linear method that combines the Lasso ($\ell_1$) and Ridge ($\ell_2$) penalties. Its optimisation problem is:

\begin{equation}
\min_{\beta} \frac{1}{2n}\lVert y - X\beta \rVert_2^2 + \lambda\left( \alpha \lVert \beta \rVert_1 + \frac{1-\alpha}{2}\lVert \beta \rVert_2^2 \right),
\label{eq:elasticnet}
\end{equation}

\noindent where $\lambda$ controls the overall regularisation and $\alpha \in [0,1]$ determines the relative weight of the $\ell_1$ and $\ell_2$ terms \citep{Zou2005}. As Eq.~\eqref{eq:elasticnet} shows, the Elastic Net is particularly useful when predictors are correlated, a scenario relevant to metals mobilised by similar hydrogeochemical processes.

Kernel ridge regression (KRR) combines Ridge regression with the kernel trick to enable non-linear modelling. Its prediction function is:

\begin{equation}
f(x) = \sum_{i=1}^{n} \alpha_i K(x_i, x),
\label{eq:krr}
\end{equation}

\noindent where $K(\cdot,\cdot)$ is a positive definite kernel and $\alpha = (K+\lambda I)^{-1}y$. The parameter $\lambda$ regularises the solution, balancing variance and bias \citep{Saunders1998}. As indicated in Eq.~\eqref{eq:krr}, polynomial kernels were adopted in this study to capture non-linear relationships between HPI and heavy metal predictors.

Having outlined the baseline regressors, the framework next integrated them within a stacked ensemble to exploit complementarities across algorithms. In stacking, predictions from base learners $\{f_j(x)\}_{j=1}^J$ are combined through a meta-learner $g(\cdot)$. The stacked predictor is expressed as:

\begin{equation}
\hat{y}(x) = g\big(f_1(x), f_2(x), \ldots, f_J(x)\big).
\label{eq:stack}
\end{equation}

\noindent Eq.~\eqref{eq:stack} formalises the two-level learning process: base learners produce preliminary predictions, and the meta-learner integrates these to improve generalisation. Base learners were selected from the baseline pool discussed above, while the meta-learner was specified as a Lasso.
The Lasso regression is a linear method with $\ell_1$-regularisation that simultaneously performs shrinkage and variable selection. Its optimisation problem is given by Eq.~\eqref{eq:lasso}:

\begin{equation}
\min_{\beta} \frac{1}{2n}\lVert y - X\beta \rVert_2^2 + \lambda \lVert \beta \rVert_1,
\label{eq:lasso}
\end{equation}

\noindent where $\lambda > 0$ controls the strength of the penalty and $\lVert \beta \rVert_1 = \sum_{j=1}^p \lVert \beta_j\rVert$ denotes the $\ell_1$ norm of the regression coefficients \citep{Tibshirani1996}. By penalising the absolute size of coefficients, Lasso tends to set some estimates exactly to zero, producing a sparse model. This property makes it particularly well-suited as a meta-learner, as it can identify and weight only the most informative base models while discarding redundant or noisy contributions.

The choice of Lasso over other baseline models for the meta-learner was motivated by several factors. First, unlike tree-based or kernel methods, Lasso produces a linear and interpretable mapping of base learners’ predictions, which is advantageous in ensemble contexts where transparency is desired. Second, the shrinkage mechanism reduces overfitting risk, especially given the relatively small sample size of environmental monitoring data from the Densu basin. Third, Lasso’s ability to handle correlated predictors is valuable in this setting, as predictions from different base models (e.g., Elastic Net and Kernel Ridge) may be collinear. Consequently, Lasso provides a balance between parsimony, interpretability, and predictive stability, making it a robust choice for stacking.

Building on this foundation, the next step in the ensemble construction involved generating out-of-fold predictions from $K$-fold partitions of the training set, which served as meta-features for the Lasso meta-learner. This procedure ensures that the meta-learner is trained on unbiased base predictions and avoids information leakage \citep{Ting1999, Wolpert1992}.

\subsubsection{Unbiased model assessment with nested cross-validation}

While ensembles enhance predictive power, careful validation is required to avoid overly optimistic results. To this end, robust model evaluation was ensured through nested cross-validation, which separates hyperparameter optimisation from performance assessment \citep{Varma2006}. In this framework, the data are split into an outer loop of $K_{\text{out}}$ folds and an inner loop of $K_{\text{in}}$ folds. For each outer fold $k=1,\ldots, K_{\text{out}}$, the training data are passed to the inner loop, where cross-validation is performed to select optimal hyperparameters $\hat{\theta}_k$. 
The model trained on the corresponding inner training set is then evaluated on the outer test fold. The overall error estimate is obtained as:

\begin{equation}
\hat{E} = \frac{1}{K_{\text{out}}} \sum_{k=1}^{K_{\text{out}}} L\big(y^{(k)}, f(x^{(k)}; \hat{\theta}_k)\big),
\label{eq:nestedcv}
\end{equation}

\noindent where $L(\cdot,\cdot)$ denotes the loss function. In this study, the root mean squared error (RMSE) was used. As Eq.~\eqref{eq:nestedcv} indicates, the error estimate represents the average performance across outer folds, thus reflecting generalisation capacity. A configuration of $K_{\text{out}}=5$ and $K_{\text{in}}=5$ was implemented, balancing bias and variance while maintaining computational feasibility.

\begin{figure}[H]
\centering
\includegraphics[width=\textwidth]{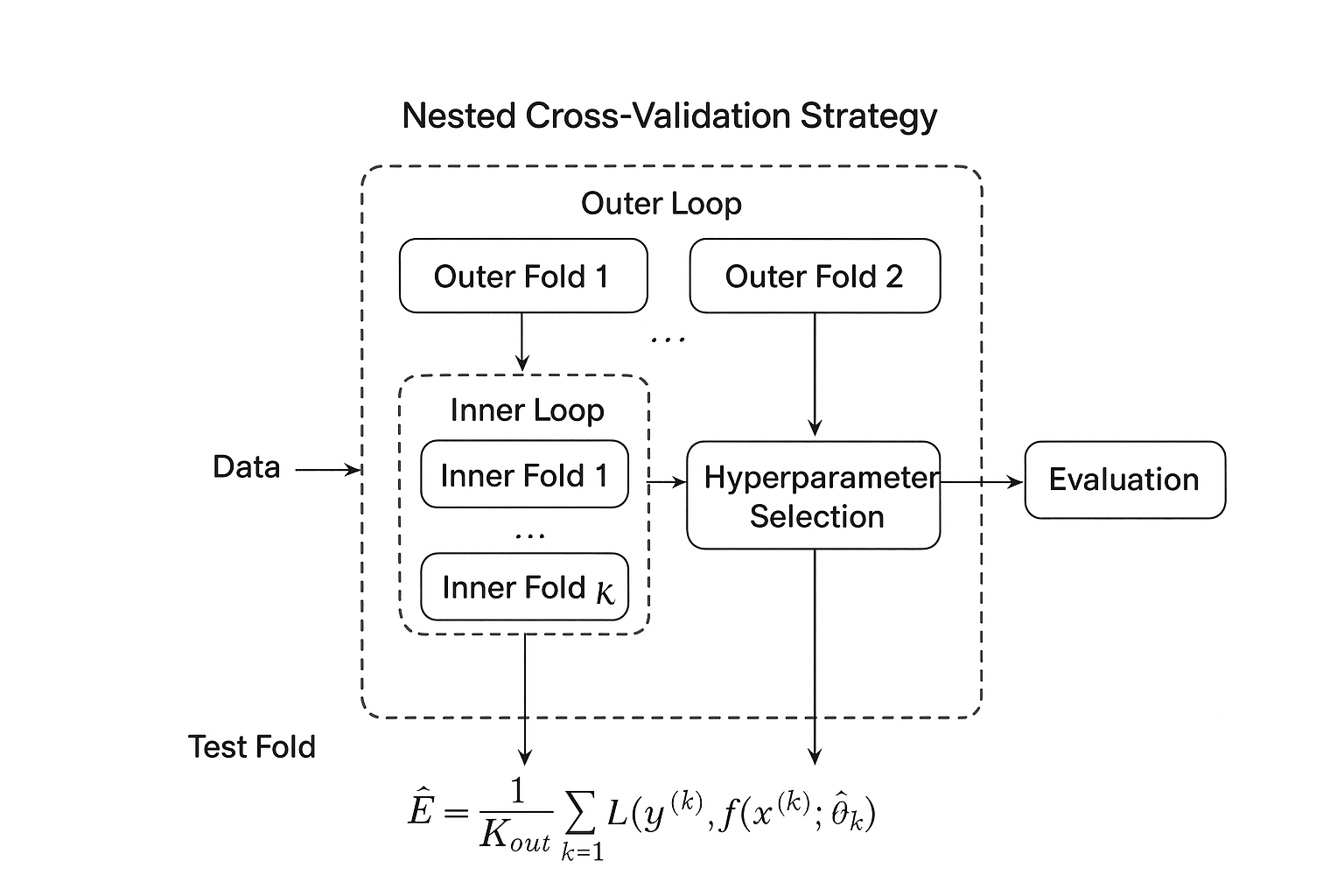}
\caption{Schematic of the nested cross-validation process. The outer loop provides an unbiased estimate of model generalisation, while the inner loop performs hyperparameter tuning.}
\label{fig:nested_cv}
\end{figure}

\noindent This nested scheme, illustrated in Fig.~\ref{fig:nested_cv}, is especially critical in ensemble modelling contexts where hyperparameter tuning can otherwise lead to optimistic bias. By decoupling optimisation from evaluation, the nested strategy ensures that reported performance estimates reflect genuine predictive capability on unseen data. Critically, all response transformations (log and Gaussian copula) were embedded within this structure: for each inner fold, transformation parameters were estimated using only the training data and applied to the corresponding validation fold, while outer fold test data remained untouched throughout, thereby preventing information leakage.

\paragraph{Implementation and hyperparameter tuning}

All modelling was implemented in Python 3.9 using the \texttt{scikit-learn} library (version 1.0.2). Hyperparameter search was performed using \texttt{GridSearchCV} within the inner folds of the nested CV, with the following ranges:

\begin{enumerate}[label=(\roman*)]
    \item SVR:  kernel = \texttt{‘rbf’},  \(\epsilon \in \{0.001, 0.01, 0.1, 0.5\}\), \\
    \(C \in \{0.1, 1, 10, 100, 1000\}\),
    \(\gamma \in \{0.001, 0.01, 0.1, 0.5, 1\}\).

    \item KRR: kernel \(\in \{\texttt{‘linear’}, \texttt{‘rbf’}, \texttt{‘polynomial’}\}\), 
    \(\alpha \in \{0.01, 0.1, 1, 10\}\), 
    \(\gamma \in \{0.001, 0.01, 0.1\}\), 
    degree \(\in \{2,3\}\).

    \item Elastic Net: 
    \(\alpha \in \{0.0001, 0.001, 0.01, 0.1, 1\}\), 
    \(\ell_1\) ratio \(\in \{0.1, 0.5, 0.7, 0.9\}\).

    \item CART: 
    max\_depth \(\in \{3,5,10,20,\text{None}\}\), 
    min\_samples\_split \(\in \{2,5,10\}\), 
    min\_samples\_leaf \(\in \{1,2,5\}\).

    \item k-NN: 
    \(k \in \{3,5,7,9,11,15\}\), 
    weights \(\in \{\texttt{‘uniform’}, \texttt{‘distance’}\}\), 
    metric \(\in \{\texttt{‘euclidean’}, \texttt{‘manhattan’}\}\).
\end{enumerate}

\noindent For the stacked ensemble, the meta‑learner (Lasso) was tuned over \(\alpha \in \{0.0001, 0.001, 0.01, 0.1, 1.0\}\). The base learners in the stacking architecture were the five individual models listed above; their hyperparameters were optimised in the inner CV loop as described. The stacking implementation followed the standard two‑level approach \citep{Wolpert1992}: out‑of‑fold predictions from the base learners were generated during the inner folds and used as features to train the meta‑learner. The final ensemble predictions were obtained by applying the trained meta‑learner to the base learners’ predictions on the outer test fold.

Given the sample size (\(n = 96\)) and the number of predictors (\(p = 6\)), the sample-to-feature ratio is \(16:1\), which is generally considered sufficient for regularised regression and ensemble methods when combined with cross-validation. All base learners incorporate explicit regularisation to constrain model complexity: SVR uses the \(C\) and \(\epsilon\) parameters; Kernel Ridge employs ridge regularisation; Elastic Net combines \(\ell_1\) and \(\ell_2\) penalties; CART is pruned via max\_depth and min\_samples\_split; and k-NN smooths predictions through the choice of \(k\). The hyperparameter search grids were deliberately limited to a few plausible values per parameter to prevent the tuning process from fitting noise. Moreover, the nested cross-validation framework provides an unbiased estimate of generalisation error; any severe overfitting would manifest as a substantial drop between inner‑fold validation performance and outer‑fold test performance. As will be shown in the Results section, the observed training–test gaps are modest, indicating that overfitting is not a major concern.

\section{Geospatial Interpolation and Mapping}\label{sec:geospatial}

\subsection{Spatial interpolation of heavy metals}\label{sec:geohvm}

The uneven spatial distribution of groundwater monitoring points within the Densu Basin limits the ability to infer basin-wide patterns of heavy metal contamination. To address this, a machine learning-based spatial interpolation framework was implemented using Random Forest (RF) regression to estimate concentrations at unsampled locations and generate spatially continuous maps for each metal.

Formally, let the observed dataset be denoted as:
\begin{equation}
\label{eq:dataset}
\mathcal{D} = \{ (\mathbf{s}_i, y_i) \}_{i=1}^n,
\end{equation}
where $\mathbf{s}_i \in \mathbb{R}^2$ represents the spatial coordinates of the $i$-th sampling location and $y_i$ the corresponding metal concentration. The interpolation task is to learn a mapping
\begin{equation}
\label{eq:function_mapping}
f: \mathbf{s} \mapsto \hat{y}(\mathbf{s}),
\end{equation}
such that $\hat{y}(\mathbf{s})$ approximates concentrations at unobserved locations.

Random Forest was selected for its ability to model non-linear spatial relationships without imposing parametric assumptions such as stationarity, which are often violated in heterogeneous environmental systems. In this framework, geographic coordinates (longitude and latitude) are used as predictor variables, allowing the model to learn spatial structure directly from the data.
Prior to model fitting, concentration values were standardised using a $z$-score transformation to ensure comparability across metals with different magnitude ranges. The RF model was configured with $n_{\text{estimators}} = 200$ trees and no restriction on tree depth.

A regular prediction grid was defined over the basin extent with a resolution of $400 \times 400$ points ($\rm \approx 200\, m$ resolution), and predictions were generated at each grid location. To ensure spatial consistency, predictions were restricted to the basin boundary using a spatial mask.
Interpolation performance was assessed using $k$-fold cross-validation on the observed data, with root mean squared error (RMSE) as the evaluation metric. The resulting interpolated concentration fields form the spatially continuous predictor inputs for subsequent HPI modelling.

\subsection{Generation of HPI spatial maps}\label{sec:geohvm1}

Following the interpolation stage, the continuous heavy metal grids were utilised as explanatory features in the prediction of the HPI. 
The HPI provides a composite measure of groundwater quality by integrating multiple heavy metal concentrations into a single index, facilitating the identification of pollution hotspots \citep{Prasad2001,Backman1998}.

Let $\mathbf{X}_g \in \mathbb{R}^{m \times p}$ denote the grid-level feature matrix, where $m$ represents the number of interpolated locations and $p$ the number of heavy metals included.
The predictive task in Eq.~\eqref{eq:hpi_prediction}, was then to estimate the HPI value at each grid point,

\begin{equation}
\label{eq:hpi_prediction}
\hat{HPI} = g(\mathbf{X}_g),
\end{equation}

\noindent  where $g(\cdot)$ corresponds to the fitted ensemble learning function.

The ensemble framework comprised both baseline regressors and higher-order meta-learners (averaged and stacked models). Model evaluation employed a nested cross-validation strategy, ensuring robust estimation of generalisation error \citep{Varma2006}.
Importantly, interpolation and mapping procedures were fully decoupled from training, with pre-fitted models applied directly to the interpolated feature grids, thereby avoiding data leakage and overoptimism.

% \paragraph{Propagation of interpolation uncertainty}

% The interpolated metal concentrations carry uncertainty arising from the limited number of sampling points and the spatial prediction model. Because HPI is a composite index that non‑linearly combines the metal concentrations (Eqs.~\ref{eq:Qi}–\ref{eq:HPI}), this uncertainty propagates into the final HPI maps. The HPI predictions presented in Section~\ref{sec:results} were evaluated against the deterministically computed HPI at the observed locations, where the HPI models were trained and validated using nested cross‑validation (Section~\ref{sec:nestedcv}); this process already accounts for the original measurement error. However, the additional uncertainty from extrapolation to unsampled grid cells is not quantified in the reported HPI prediction errors. Hence, while the maps provide useful spatial patterns, the absolute values should be interpreted with caution, and future work may incorporate formal uncertainty propagation (e.g., via sequential Gaussian simulation or Bayesian hierarchical modelling).

\section{Model Performance Metrics}\label{sec:metrics}

The performance of the predictive framework was evaluated using a comprehensive suite of statistical measures spanning error magnitude, goodness-of-fit, and agreement statistics. Employing multiple metrics reduces reliance on a single criterion, thereby providing a balanced assessment of predictive skill in groundwater quality applications characterised by spatial heterogeneity and skewed distributions \citep{Gupta2009,Willmott2005}.

\subsection{Error-based metrics} \label{sec: err_metrics}
The RMSE is widely adopted as a measure of predictive accuracy as given Eq.\eqref{eq:rmse}:

\begin{equation}
\label{eq:rmse}
\text{RMSE} = \sqrt{\frac{1}{n}\sum_{i=1}^{n} \left(y_i - \hat{y}_i\right)^2},
\end{equation}

\noindent where $y_i$ and $\hat{y}_i$ are the observed and predicted values, respectively, and $n$ is the sample size. RMSE penalises larger deviations more heavily, making it sensitive to outliers.

To mitigate this sensitivity, the mean absolute error (MAE) was also employed (see Eq.~\eqref{eq:mae}):

\begin{equation}
\label{eq:mae}
\text{MAE} = \frac{1}{n}\sum_{i=1}^{n} \lvert y_i - \hat{y}_i\rvert.
\end{equation}

\noindent  Unlike RMSE, MAE weights all deviations equally, providing a robust alternative when extreme residuals occur.

The median absolute error (MedAE) in Eq.~\eqref{eq:mae} further enhances robustness against outliers by considering the median rather than the mean:

\begin{equation}
\label{eq:med}
\text{MedAE} = \text{median}\left( \lvert y_i - \hat{y}_i \rvert \right).
\end{equation}

Finally, the maximum error (MaxError) quantifies the single worst-case deviation as expressed in Eq.~\eqref{eq:maxerr}:

\begin{equation}
\label{eq:maxerr}
\text{MaxError} = \max_{i} \lvert y_i - \hat{y}_i \rvert,
\end{equation}

\noindent   which is particularly relevant in groundwater risk mapping where isolated but extreme mispredictions can mask localised contamination hotspots.

\subsection{Goodness-of-fit and information-theoretic criteria} \label{sec: gof_metrics}

The explanatory power of the models was assessed through the coefficient of determination as given in in Eq.~\eqref{eq:r2}:

\begin{equation}
\label{eq:r2}
R^2 = 1 - \frac{\sum_{i=1}^{n} (y_i - \hat{y}_i)^2}{\sum_{i=1}^{n} (y_i - \bar{y})^2},
\end{equation}

\noindent   where $\bar{y}$ denotes the mean of observed values. Although intuitive, $R^2$ may overstate performance in non-linear contexts. The adjusted $R^2$ corrects for the number of predictors ($p$) relative to sample size ($n$):

\begin{equation}
\label{eq:r2_adjust}
R^2_{\text{adj}} = 1 - \left(1 - R^2\right) \frac{n - 1}{n - p - 1}.
\end{equation}

To explicitly penalise complexity, information-theoretic criteria were employed. The Akaike Information Criterion (AIC) is defined as:

\begin{equation}
\label{eq:aic}
\text{AIC} = n \ln\left(\frac{\text{SSE}}{n}\right) + 2k,
\end{equation}

\noindent  where $\text{SSE} = \sum_{i=1}^{n} (y_i - \hat{y}_i)^2$ is the sum of squared errors and $k$ is the number of estimated parameters. The Bayesian Information Criterion (BIC) introduces a stronger penalty on model complexity:

\begin{equation}
\label{eq:bic}
\text{BIC} = n \ln\left(\frac{\text{SSE}}{n}\right) + k \ln(n).
\end{equation}

\noindent  Both criteria are essential when comparing models with different degrees of flexibility, as in the present ensemble framework \citep{Burnham2004}.

\subsection{Agreement and distributional measures} \label{sec: agg_metrics}

The concordance correlation coefficient (CCC) integrates accuracy and precision by evaluating deviation from the line of perfect concordance ($y = \hat{y}$):

\begin{equation}
\label{eq:ccc}
\text{CCC} = \frac{2 \, \text{Cov}(y, \hat{y})}{\sigma_y^2 + \sigma_{\hat{y}}^2 + \left(\mu_y - \mu_{\hat{y}}\right)^2},
\end{equation}

\noindent  where $\text{Cov}(y, \hat{y})$ is the covariance, $\mu_y$ and $\mu_{\hat{y}}$ are means, and $\sigma_y^2$, $\sigma_{\hat{y}}^2$ are variances of observed and predicted values, respectively \citep{Lin1989}. CCC values close to $1$ indicate both strong correlation and minimal bias.

The Kolmogorov–Smirnov (KS) test was used to compare the distribution of residuals with that of a reference normal distribution:

\begin{equation}
\label{eq:ks}
D_{n} = \sup_x \vert F_n(x) - F(x) \vert,
\end{equation}

\noindent  where  $F_{n}(x)$ is the empirical CDF of residuals and $F(x)$ is the theoretical CDF under normality. A significant statistic indicates structural deviations in residual distribution.

The reduced chi-squared statistic normalises error relative to variance and degrees of freedom:

\begin{equation}
\label{eq:chi}
\chi^2_\nu = \frac{1}{\nu} \sum_{i=1}^{n} \frac{\left(y_i - \hat{y}_i\right)^2}{\sigma^2},
\end{equation}

\noindent  where $\nu = n - k$ is the degrees of freedom and $\sigma^2$ is the residual variance. A value close to 1 suggests a model consistent with the assumed error structure.

The integration of these complementary metrics enabled model evaluation across multiple dimensions of performance. Error-based measures such as RMSE and MAE quantified aggregate predictive deviations, whereas MaxError provided insight into the magnitude of worst-case local departures between predictions and observations. Information criteria, namely AIC and BIC, served to assess the trade-off between model fit and complexity, offering a formal basis for comparing competing learners. Agreement and distributional metrics, including CCC and the KS test, extended this evaluation by determining whether predictions aligned with observed values not only in terms of central tendency but also in their underlying distributional structure. Together, these metrics ensured that model assessment was sensitive both to average predictive accuracy and to robustness under heterogeneous spatial conditions.

%%% 
\section{Results and Discussion} \label{sec:results}

\subsection{Analysis of prediction errors and residuals}\label{sec}
\begin{figure}[H]
\centering
\includegraphics[width=0.9\textwidth]{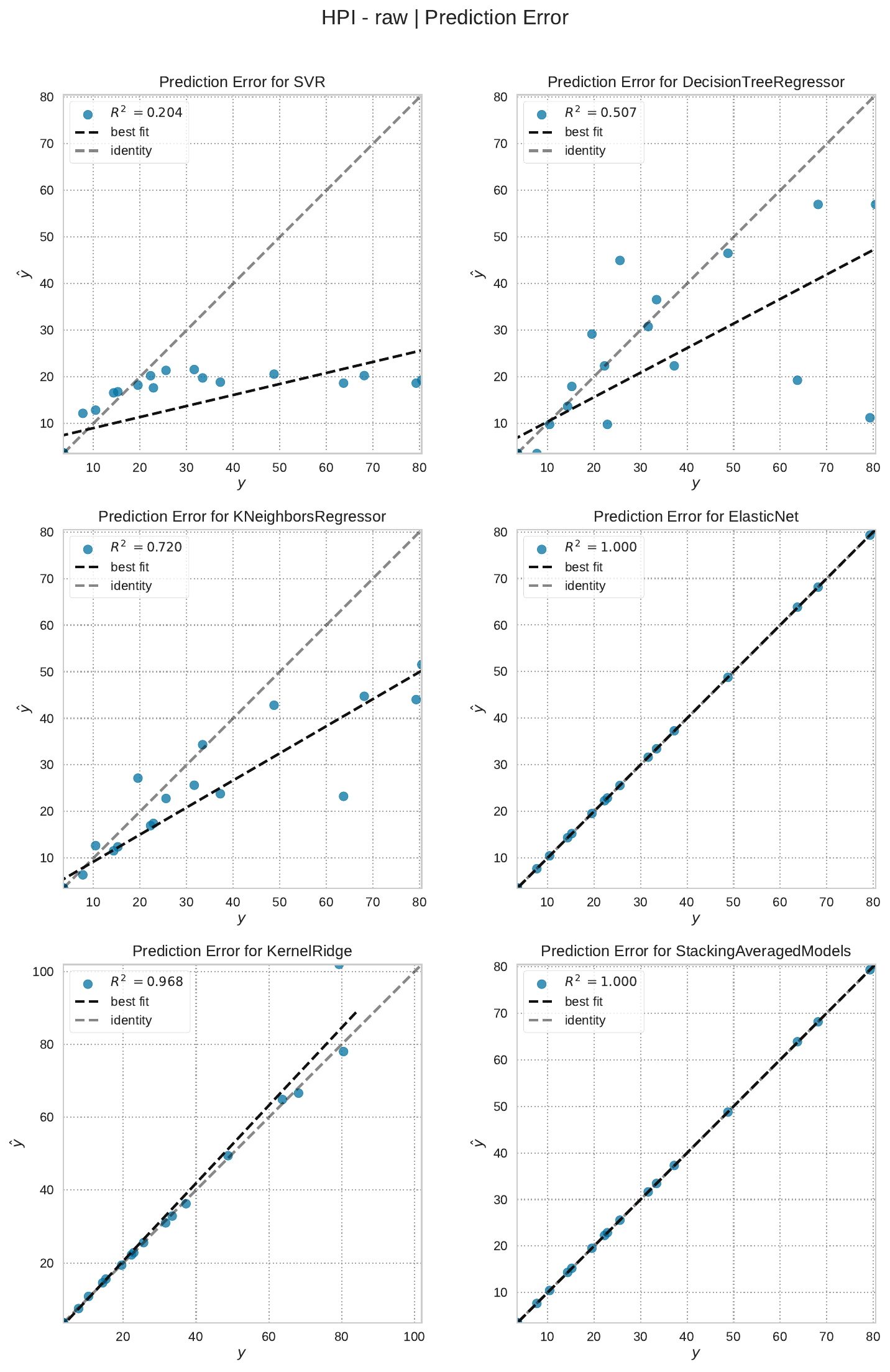}
\caption{Prediction error plots for HPI on the raw scale. Each panel compares observed and predicted values for five base learners (SVR, CART, k-NN, Elastic Net, and Kernel Ridge) and the stacked ensemble (Lasso meta-learner). The identity line and fitted regression line are shown for reference, with the corresponding $R^2$ statistic reported for the test set in each case.}
\label{fig:PredictionError_raw}
\end{figure}

\begin{figure}[H]
\centering
\includegraphics[width=0.9\textwidth]{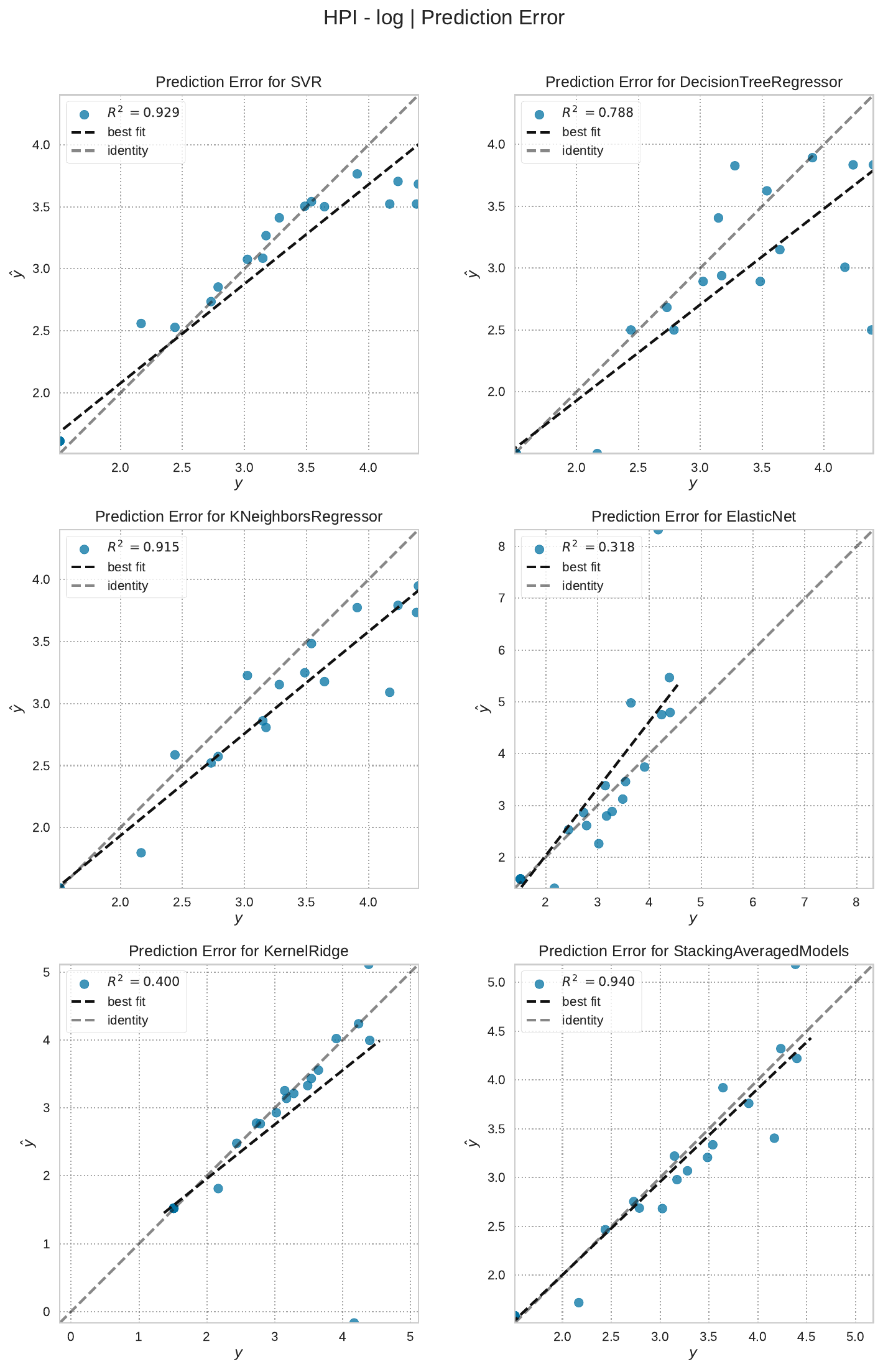}
\caption{Prediction error plots for log-transformed HPI. Panels show observed versus predicted values for five base learners and the stacked ensemble, with identity and fitted regression lines. The log transformation reduces skewness and compresses extreme values, resulting in more stable error patterns across algorithms.}
\label{fig:PredictionError_log}
\end{figure}

\begin{figure}[H]
\centering
\includegraphics[width=0.9\textwidth]{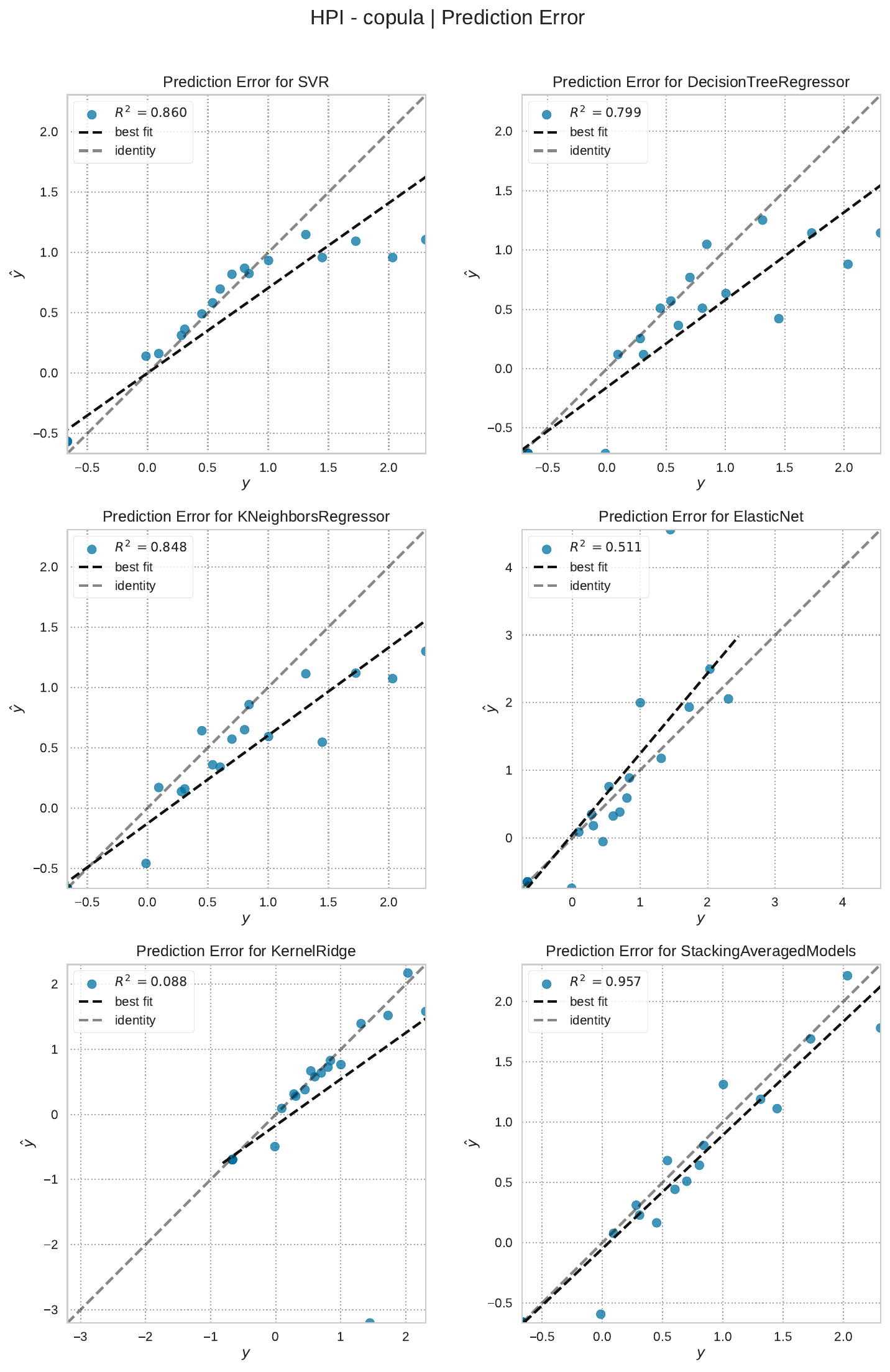}
\caption{Prediction error plots for HPI after Gaussian-copula transformation. Each panel presents observed versus predicted values for five base learners and the stacked ensemble, together with the identity and fitted regression lines. The copula transformation maps marginal distributions to a Gaussian scale while preserving dependence among variables, providing a basis for more consistent comparison of model outputs across algorithms.}
\label{fig:PredictionError_copula}
\end{figure}

\begin{figure}[H]
\centering
\includegraphics[width=\textwidth]{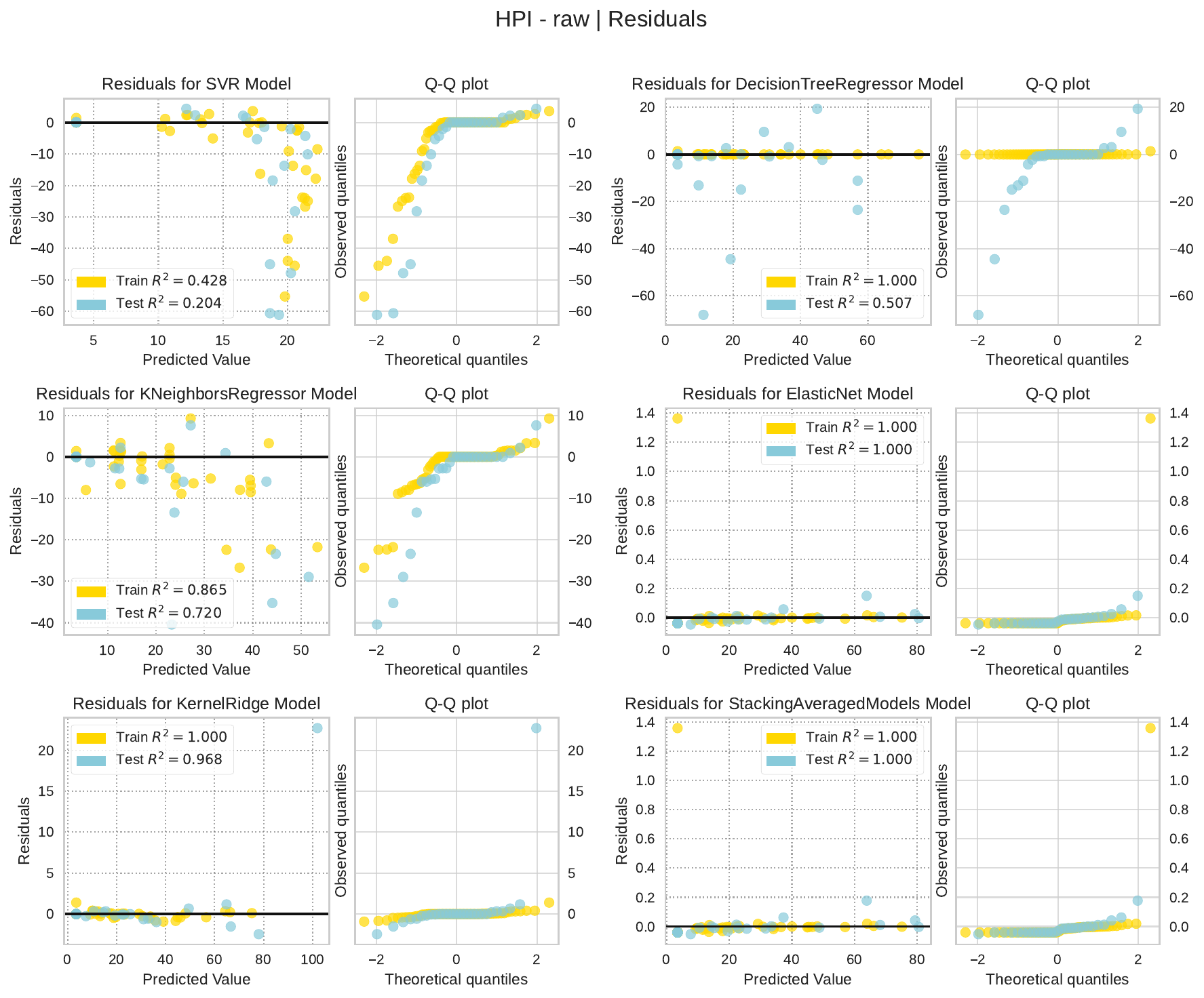}
\caption{Residual diagnostics for HPI (raw): residuals vs predicted (train and test) and Q–Q plots. Panels show heteroscedasticity and non-normal residual structure for many learners when the raw HPI is modelled; stacked predictions display unusually low residual variance on both training and test sets}
\label{fig:ResidualsPlot_raw}
\end{figure}

\begin{figure}[H]
\centering
\includegraphics[width=\textwidth]{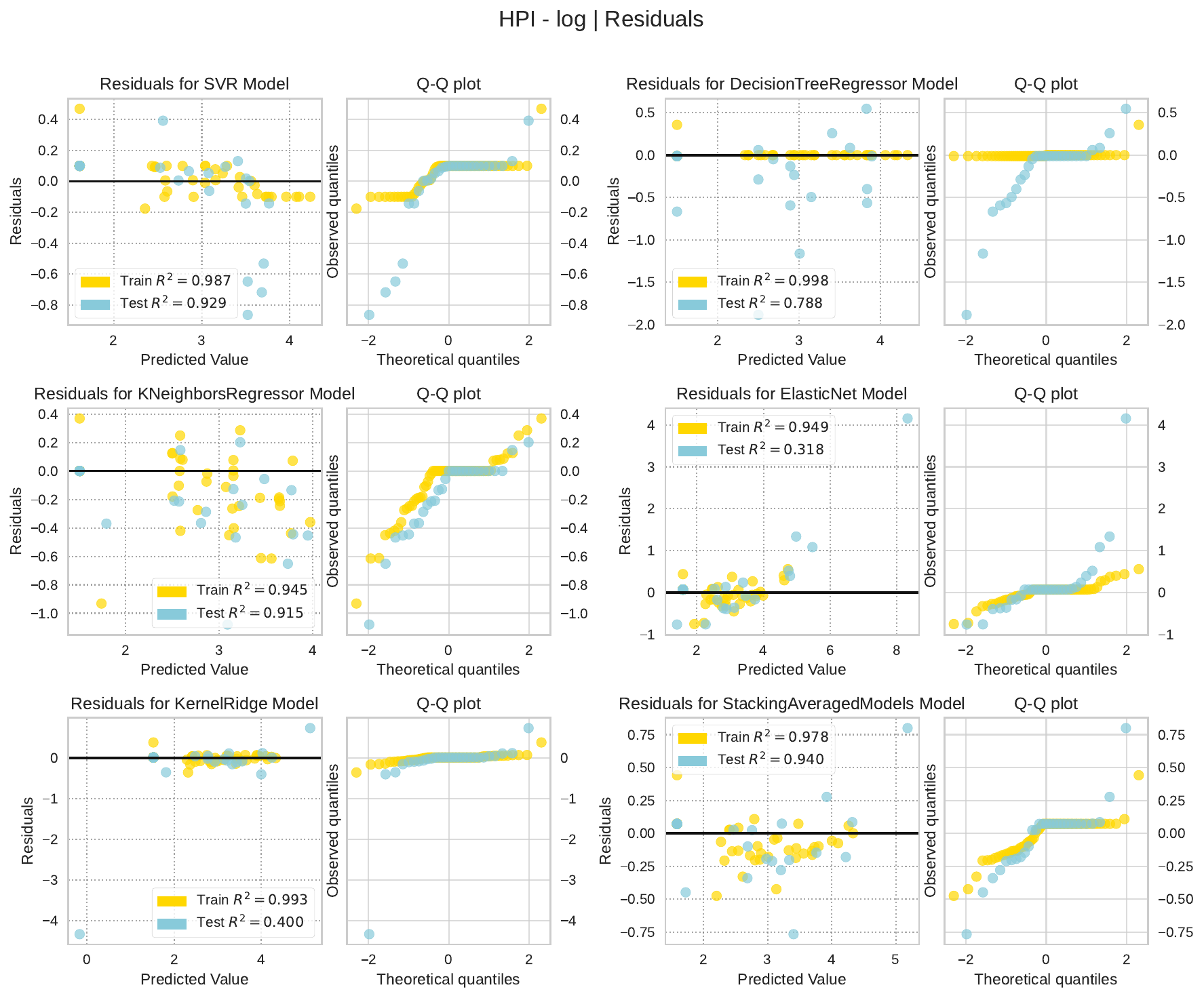}
\caption{Residual diagnostics for log-transformed HPI. Panels show residuals against predicted values for training and test sets, together with corresponding Q–Q plots. The log transformation moderates variance heterogeneity and shifts residual distributions closer to normality for several learners, although some models continue to exhibit patterns suggestive of specification issues.}
\label{fig:ResidualsPlot_log}
\end{figure}

\begin{figure}[H]
\centering
\includegraphics[width=\textwidth]{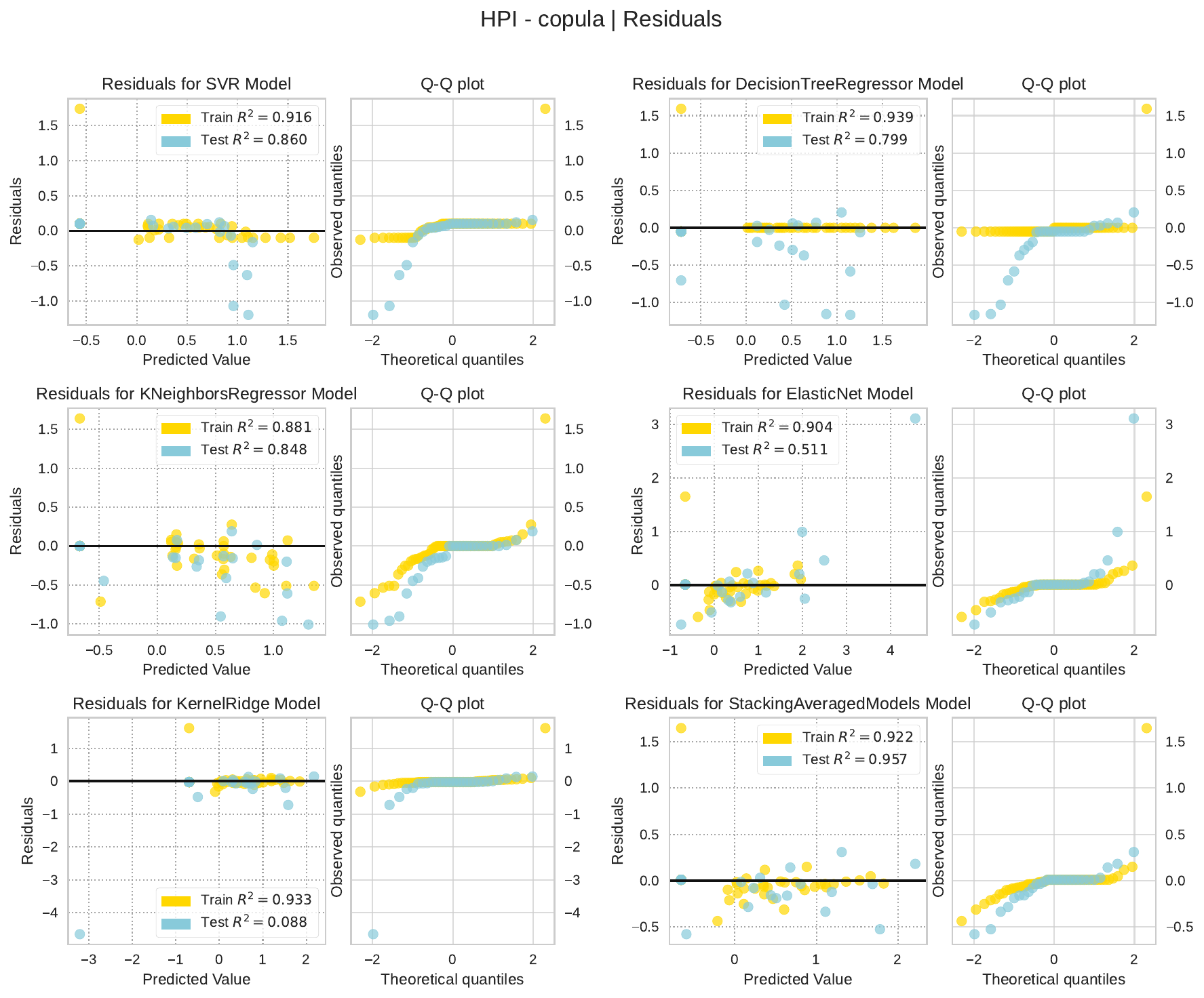}
\caption{Residual diagnostics for HPI (Gaussian copula): residuals vs predicted (train and test) and Q–Q plots. Residuals from copula-transformed models show the most homoscedastic and near-normal behaviour across learners, and the stacked ensemble produces well-behaved residuals and Q-Q alignment compared with raw and log cases. These diagnostics support the use of copula margins for stabilising the response distribution prior to regression.}
\label{fig:ResidualsPlot_copula}
\end{figure}

Figs.~\ref{fig:PredictionError_raw}--\ref{fig:PredictionError_copula} summarise the prediction-error relationships between observed and predicted HPI values under three target representations (raw, log-transformed, and Gaussian copula), while Figs.~\ref{fig:ResidualsPlot_raw}--\ref{fig:ResidualsPlot_copula} present the corresponding residual and Q–Q diagnostics. 

On the raw scale (Figs.~\ref{fig:PredictionError_raw}), the Elastic Net and stacked ensemble models returned near-perfect predictions with reported coefficients of determination approaching unity ($R^2 \approx 1.00$), and extremely low RMSE values. 
While superficially impressive, such performance warrants caution. 
The near-unity fit for both training and test partitions suggests potential information leakage or overly optimistic model behaviour, a known risk in ensemble learning if cross-validation protocols are inadvertently misapplied.
By contrast, kernel ridge regression ($R^2 \approx 0.97$, RMSE $\approx 0.23$) and k-NN ($R^2 \approx 0.72$, RMSE $\approx 0.78$) demonstrated more moderate fits, whereas support vector regression ($R^2 \approx 0.20$, RMSE $\approx 2.12$) underperformed, producing substantial scatter around the identity line. These results illustrate how untransformed HPI values amplify heteroscedasticity and skewness, disproportionately influencing certain algorithms and obscuring realistic predictive structure.

The log transformation (Fig.~\ref{fig:PredictionError_log}) alleviated this imbalance by compressing extreme HPI values, thereby improving performance for several learners. SVR achieved $R^2 \approx 0.93$ (RMSE $\approx 0.18$), while k-NN performed comparably ($R^2 \approx 0.92$, RMSE $\approx 0.20$). 
The stacked ensemble retained strong predictive power ($R^2 \approx 0.94$, RMSE $\approx 0.16$), though without the implausible perfection of the raw-scale models.
Elastic Net, however, degraded substantially ($R^2 \approx 0.32$, RMSE $\approx 0.51$), indicating sensitivity to the reduced variance in the log-transformed response. Kernel ridge regression also suffered ($R^2 \approx 0.40$, RMSE $\approx 0.42$), suggesting that non-linear kernel mapping was less effective when tail values were compressed. These outcomes confirm that the log transformation mitigates skewness-induced overfitting for some learners while attenuating performance for others, particularly those dependent on linear shrinkage or kernel-based regularisation.

%% Prediction Error plot

The Gaussian copula transformation (Fig.~\ref{fig:PredictionError_copula}) yielded the most balanced and statistically robust results.
Here, the stacked ensemble again achieved the strongest performance ($R^2 \approx 0.96$, RMSE $\approx 0.19$), while SVM ($R^2 \approx 0.86$, RMSE $\approx 0.25$) and k-NN ($R^2 \approx 0.85$, RMSE $\approx 0.26$) also maintained high predictive accuracy. Decision trees (CART) performed moderately well ($R^2 \approx 0.80$, RMSE $\approx 0.30$), whereas Elastic Net ($R^2 \approx 0.51$, RMSE $\approx 0.39$) and kernel ridge regression ($R^2 \approx 0.09$, RMSE $\approx 0.69$) struggled to leverage the copula margins effectively. 

Crucially, residual plots (Figs.~\ref{fig:ResidualsPlot_raw}--\ref{fig:ResidualsPlot_copula}) corroborate these patterns. For raw HPI, residuals of Elastic Net and the stacked model cluster unrealistically close to zero for both training and test partitions, an artefact suggestive of leakage.
In contrast, log and copula transformations produced residual distributions that were more homoscedastic and symmetrically centred around zero, with Q–Q plots indicating closer adherence to Gaussian assumptions, particularly under the copula framework.

%% Residual plot

A critical implication of these diagnostics is that models which appear to achieve near-perfect predictive fit on the raw HPI scale may, in practice, obscure the true variability of heavy-metal contamination. Overly smooth or overconfident predictions risk downplaying the small-scale heterogeneity that is often characteristic of groundwater systems in heterogeneous geological settings \citep{macdonald2012quantitative,Edmunds2012}.
In contrast, the copula-based models generated error structures that were both statistically credible and consistent with the expectation of spatially variable contaminant distribution. Similar benefits of distributional transformations have been demonstrated in water quality prediction studies, where copula frameworks preserved dependence among multiple hydrochemical indicators while stabilising variance \citep{graler2014modelling}. By improving residual behaviour and enhancing model generalisability, the copula transformation therefore provides a more defensible basis for predictive HPI mapping compared to raw or log scales.

Concerns regarding overfitting with complex models on limited samples are mitigated by several aspects of our framework. The nested cross-validation protocol provides unbiased generalisation estimates by completely separating hyperparameter optimisation from performance assessment. The ensemble architecture incorporates multiple regularisation mechanisms across base learners and the Lasso meta-learner, while the consistent performance across transformations and the spatial coherence of predictions (discussed in Section~\ref{sec:sp_HPImaps}) suggest that models are capturing genuine hydrogeochemical structure rather than memorising noise. The modest gap between training and test $R^2$ values in Figs.~\ref{fig:ResidualsPlot_raw}--\ref{fig:ResidualsPlot_copula} further supports the stability of our approach.

\subsection{Spatial distribution of heavy metals} \label{sec:sp_hvymaps}

\begin{figure}[H]
\centering
\includegraphics[width=0.9\textwidth]{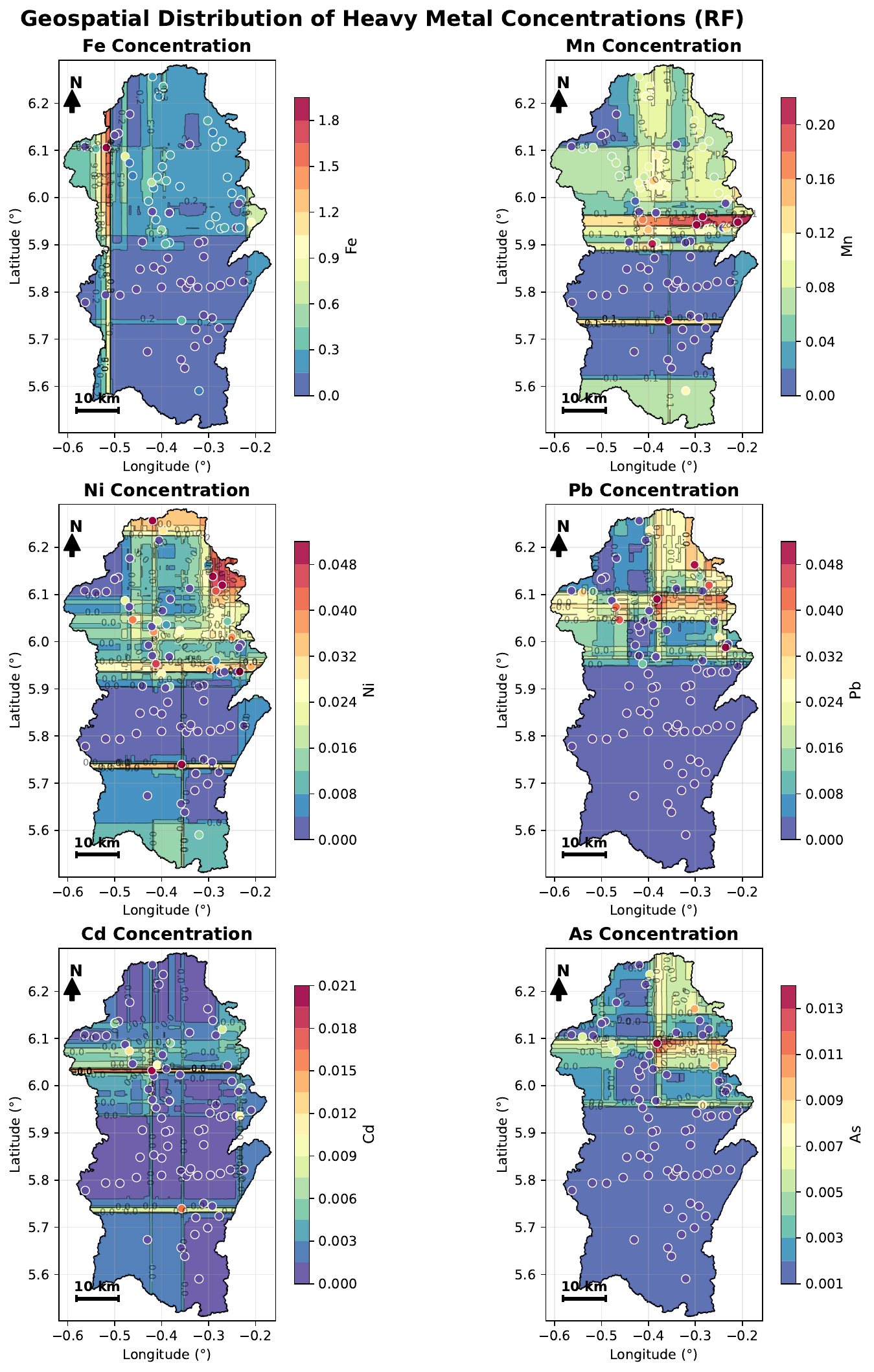}
\caption{Spatial distribution of RF-interpolated heavy-metal concentrations (mg L$^{-1}$) across the Densu Basin. Fe and Mn dominate the profiles, while Ni and Pb occur in patchy, localised zones. Cd and As remain near detection limits across most of the basin. Circles denote sampling locations.}
\label{fig:spatial_map_rf}
\end{figure}

The RF–interpolated maps of heavy metal concentrations across the Densu Basin (Fig.~\ref{fig:spatial_map_rf}) reveal marked spatial heterogeneity that is closely tied to the region's geology, redox conditions, and land use. Iron dominates the concentration profiles, with hotspot values exceeding $0.33$~mgL$^{-1}$ in the central–northern parts of the basin. These elevated Fe levels are consistent with mobilisation from Fe-bearing minerals and reductive dissolution of oxyhydroxide phases, a behaviour frequently reported in tropical alluvial aquifers \citep{Smedley2002}.Manganese exhibits a comparable, though less intense, distribution pattern, attaining concentrations of around $0.10$~mgL$^{-1}$ in discrete zones that frequently overlap with elevated Fe. This spatial association suggests common geochemical controls, since both elements are strongly influenced by redox conditions and mineral dissolution processes. The underlying Birimian metasediments and Tarkwaian sandstones contain abundant ferromagnesian minerals, which, upon weathering, release Fe and Mn into the groundwater \citep{akurugu2022groundwater}. In settings with shallow water tables and high organic‑matter content, especially agricultural areas--reducing conditions are established, which favour the reductive dissolution of Fe–Mn oxyhydroxides \citep{Saeed2023, Zhai2022}. This mechanism liberates Fe$^{2+}$ and Mn$^{2+}$ into solution and can simultaneously mobilise trace elements previously adsorbed to the oxyhydroxide surfaces.

Nickel and lead exhibit much lower overall values, with maximum interpolated concentrations of approximately $0.02$~mgL$^{-1}$ and $0.014$~mgL$^{-1}$, respectively. Their distributions are patchier, occurring in scattered localised zones rather than basin‑wide. These anomalies likely reflect anthropogenic inputs such as urban discharges or artisanal mining activities, which have been previously noted in the basin \citep{Armah2014}. However, the co‑occurrence of Ni with Fe–Mn hotspots in some areas also suggests a natural co‑mobilisation pathway during reductive dissolution. Cadmium and arsenic are the least abundant, with concentrations close to baseline detection levels ($0.003$–$0.004$~mgL$^{-1}$) across most of the basin. Their very low and spatially uniform pattern suggests that these elements are either absent in significant geogenic reservoirs in the Densu aquifer system, or their mobilisation is limited under the prevailing pH and redox conditions. Consequently, the Fe–Mn dominated cluster (Cluster--$1$ in Fig.~\ref{fig:clusterPlot}) corresponds to zones where reducing conditions and mineral weathering are most active, while the background cluster (Cluster--$0$) represents areas with oxidising conditions or where Fe–Mn oxyhydroxides remain stable. Overall, the metal maps show a basin dominated by Fe and Mn, with secondary contributions from Ni and Pb in isolated areas. Cadmium and As appear largely negligible at the basin scale, though this does not preclude localised enrichment below the current sampling resolution.

\paragraph{Interpolation performance and uncertainty.}
The accuracy of the RF interpolation was evaluated using cross-validation RMSE for each metal. The results indicate varying levels of predictive accuracy across metals, with higher errors observed for Fe (0.188 mg L$^{-1}$) and Mn (0.058 mg L$^{-1}$), and substantially lower errors for trace metals such as Ni (0.016 mg L$^{-1}$), Pb (0.014 mg L$^{-1}$), Cd (0.004 mg L$^{-1}$), and As (0.003 mg L$^{-1}$). This pattern reflects differences in spatial variability and concentration ranges, with Fe and Mn exhibiting greater heterogeneity across the basin.

The interpolated metal concentrations carry uncertainty arising from the limited number of sampling points and the spatial prediction model. Because HPI is a composite index that non-linearly combines the metal concentrations (Eqs.~\ref{eq:subindex}--\ref{eq:agg}), this uncertainty propagates into the final HPI maps. Model performance at observed locations is evaluated using cross-validation and an independent test split; however, this assessment does not capture the additional uncertainty associated with spatial prediction at unsampled grid locations. Consequently, while the maps provide useful insight into spatial patterns of groundwater quality, the absolute HPI values should be interpreted with caution. Formal uncertainty propagation (e.g., via geostatistical simulation or Bayesian hierarchical modelling) remains an important direction for future work.

\subsection{Spatial distribution of predicted HPI maps} \label{sec:sp_HPImaps}

\begin{figure}[H]
\centering
\includegraphics[width=0.90\textwidth]{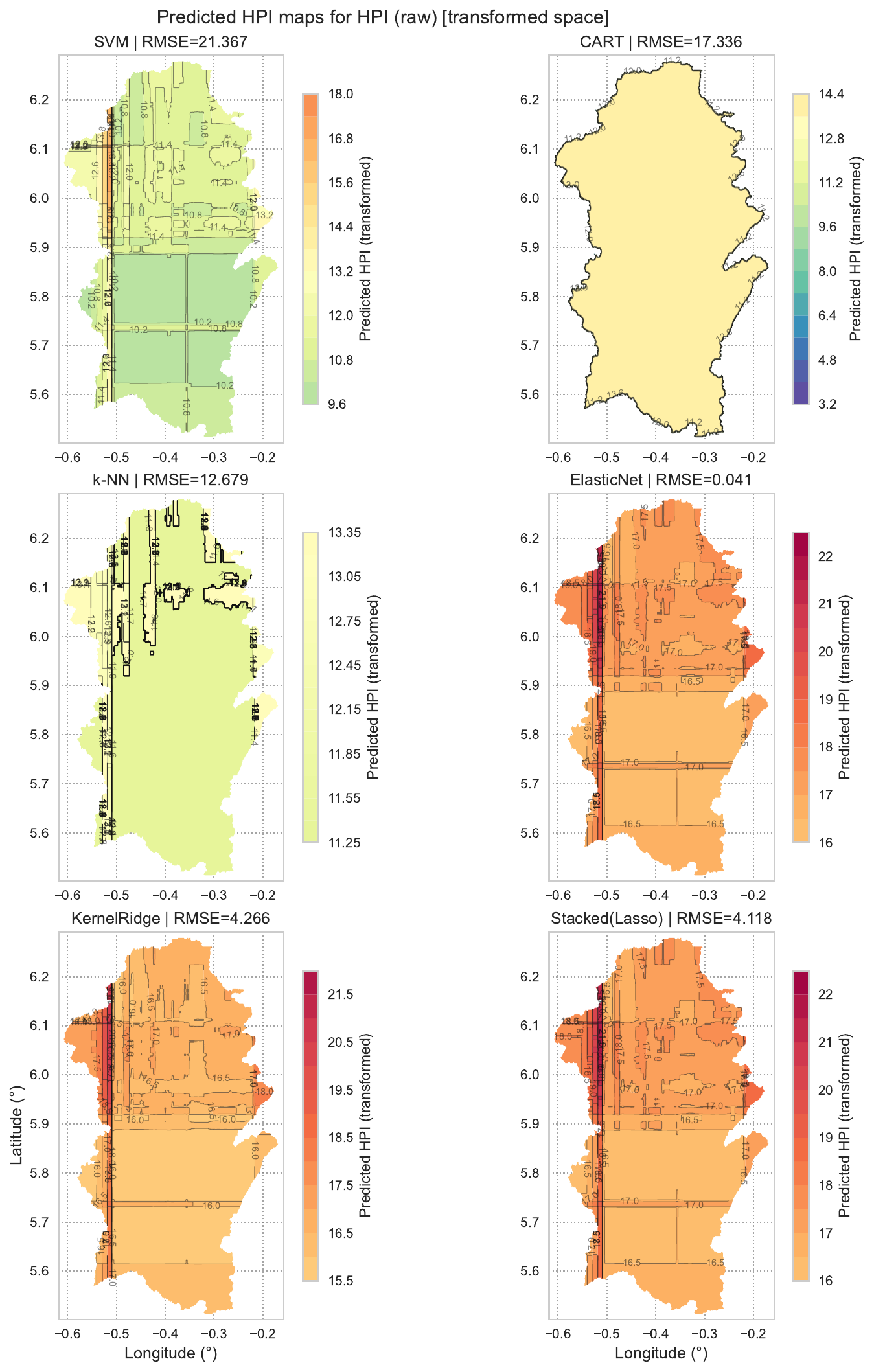}
\caption{Predicted HPI spatial maps obtained by applying pre-trained baseline and ensemble models to the RF-interpolated heavy-metal grids, using raw HPI as the response. Panels are labelled with model type and RMSE. Most models, including SVM, k-NN, Kernel Ridge, Elastic Net, and the stacked ensemble, consistently resolve a hotspot in the northwest of the Densu basin. By contrast, CART yields an overly uniform prediction, indicating limited capacity to capture spatial heterogeneity.}
\label{fig:HPI_maps_HPI_raw}
\end{figure}

The HPI maps generated using the pre-trained models with the raw form of the target variable (refer to Fig.~\ref{fig:HPI_maps_HPI_raw}) highlight model-dependent differences in predictive structure.  
Among the baseline learners, SVM produced widespread moderate predictions with RMSE values $>21$, indicating limited fit to the underlying spatial variability.  
By contrast, Kernel Ridge regression achieved markedly lower RMSE values ($\sim 4.3$), with maps showing higher predicted HPI values ($>20$) in the north-western sector of the basin.  
The stacked ensemble further improved predictive stability, with an RMSE of $\sim 4.1$, and captured both the elevated northwest region and a secondary hotspot near the central corridor.  
CART, on the other hand, produced near-homogeneous surfaces with RMSE $\sim 17$, indicating underfitting and poor spatial resolution of local anomalies. k-NN offered more localised predictions (RMSE $\sim 12.7$) but closely mirrored the sample distribution, leading to striping patterns where monitoring points were sparse.
Elastic Net achieved an extremely low training error (RMSE $\sim 0.04$) and generated spatial predictions that were consistent with both Kernel Ridge and the stacked ensemble. It highlighted the same north-western hotspot as SVM and k-NN, while also capturing a secondary elevated corridor through the central basin. The maximum predicted HPI values $> 22$, which is comparable to the ranges observed in Kernel Ridge and ensemble outputs, indicating that Elastic Net captured genuine spatial variability rather than producing unstable artefacts.
Collectively, these results demonstrate that kernel-based models (Kernel Ridge, SVM) and the stacked ensemble provide the most reliable representations of HPI gradients in the basin, while CART and k-NN remain constrained by oversmoothing and local sampling density, respectively.

%%% HPI -->> Raw

\begin{figure}[H]
\centering
\includegraphics[width=0.90\textwidth]{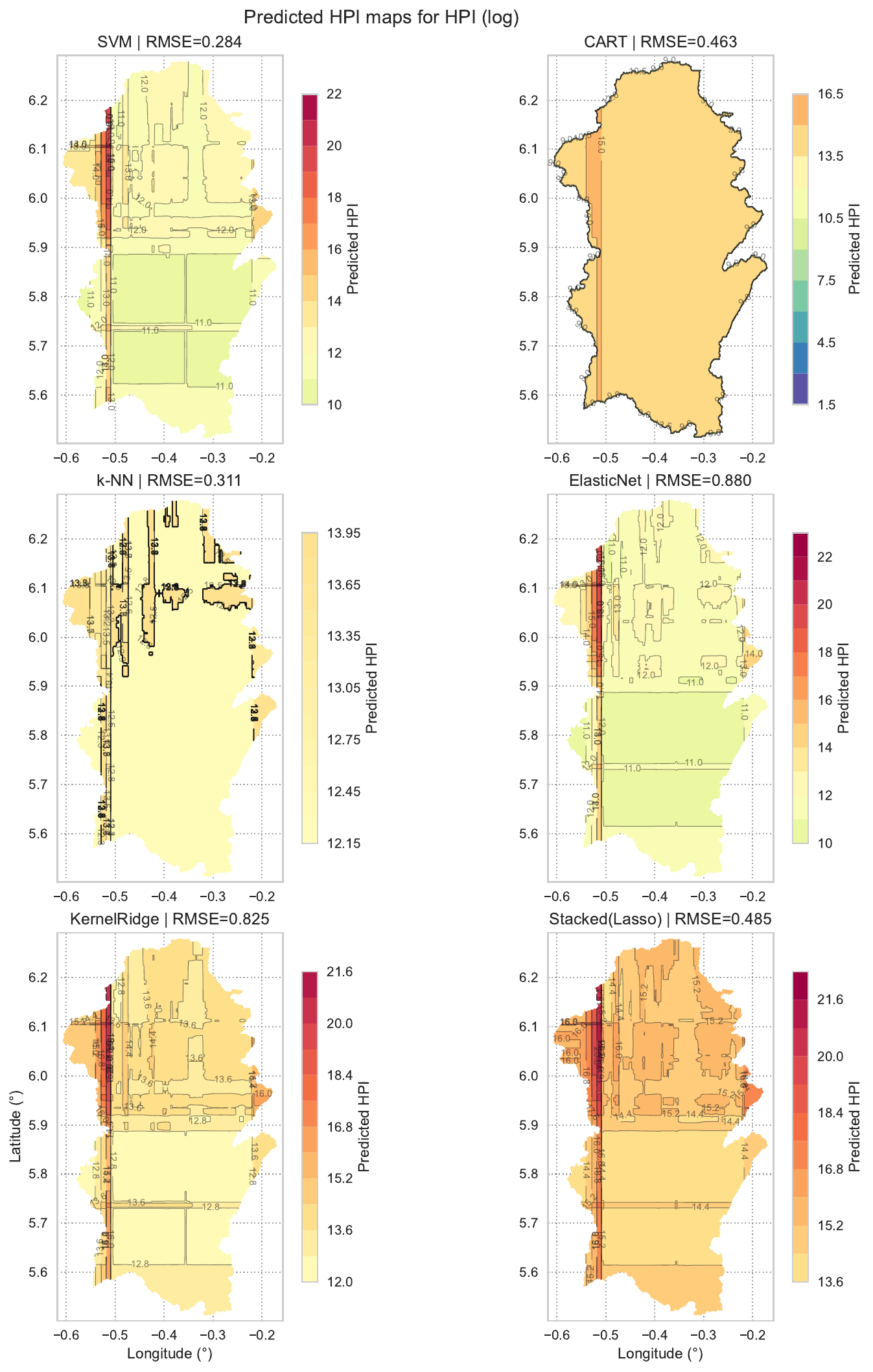}
\caption{
Predicted HPI spatial maps obtained by applying pre-trained baseline and ensemble models to RF-interpolated heavy-metal grids, with models trained using log-transformed HPI as the response. The log transformation improved predictive stability, yielding consistent hotspots across models and reducing irregular spatial variability compared to raw-scale predictions.}
\label{fig:HPI_maps_HPI_log_to_raw}
\end{figure}

When HPI was modelled on the log-transformed scale of the pre-trained nested cross-validated models (Fig.~\ref{fig:HPI_maps_HPI_log_to_raw}), prediction errors decreased markedly across most learners. SVM achieved RMSE values of $\sim 0.28$, while Kernel Ridge and k-NN produced comparably low errors ($0.31$--$0.83$). The stacked ensemble also performed strongly (RMSE $\sim 0.49$). These improvements are substantial when compared to the raw-scale predictions, underscoring the effectiveness of log transformation in stabilising skewed HPI distributions. The spatial maps reinforce this finding: predicted surfaces are smoother, less influenced by local sampling density, and consistently highlight hotspots in the northwest and upper-central sectors of the basin. 
CART continued to yield near-uniform predictions (RMSE $\sim 0.46$), suggesting limited ability to capture basin-scale heterogeneity. Elastic Net, though reasonably accurate (RMSE $\sim 0.88$), exhibited localised over-prediction in parts of the central basin. Overall,  the log-transformed models converged towards more consistent spatial structures across algorithms, indicating that the transformation reduced model-specific biases and enhanced the robustness of basin-wide predictions.

\begin{figure}[H]
\centering
\includegraphics[width=0.9\textwidth]{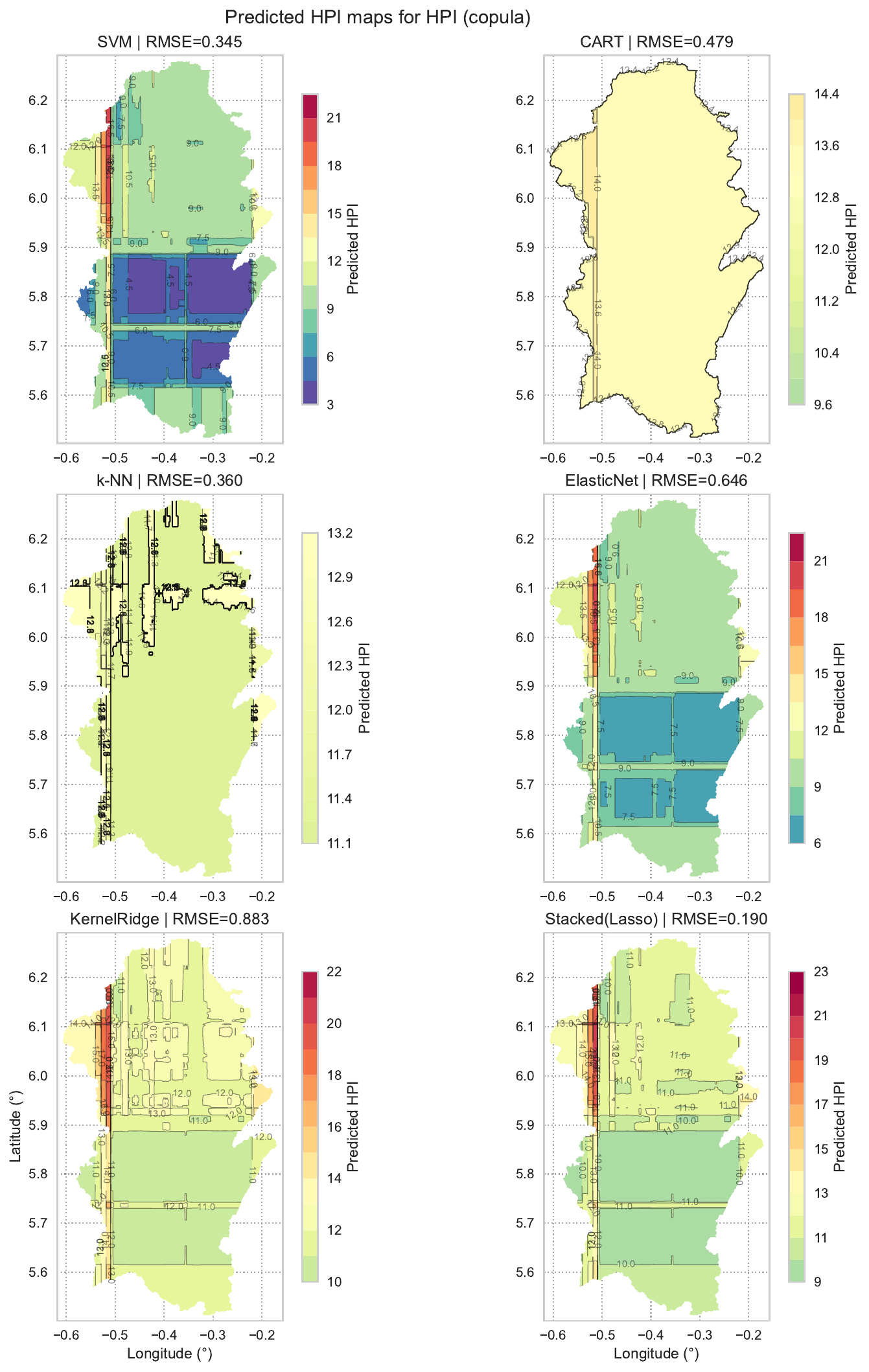}
\caption{Spatial distribution of predicted Heavy Metal Pollution Index (HPI) values across the Densu Basin under the Gaussian copula transformation, comparing six machine learning algorithms (SVM, CART, k-NN, Elastic Net, Kernel Ridge, and Stacked Lasso). The maps highlight basin-wide heterogeneity in predicted groundwater quality, with model-specific variations in spatial patterning and prediction accuracy (RMSE values reported for each model).}
\label{fig:HPI_maps_HPI_copula_to_raw}
\end{figure}

Building on the improvements observed with the log-transformed framework, the Gaussian copula transformation provided an additional layer of refinement by simultaneously normalising marginal distributions and preserving dependence structures \citep{Krupskii2018}.
Fig.~\ref{fig:HPI_maps_HPI_copula_to_raw} illustrates the predicted HPI maps derived from six representative algorithms under this transformation. Notably, the stacked ensemble (Lasso meta-learner) achieved the lowest RMSE ($\sim 0.19$), producing a coherent basin-wide prediction surface that delineated zones of moderate to high pollution in the central and north-western corridors. These outputs align with previously identified hydrogeochemical hotspots where Fe and Mn mobilisation is pronounced \citep{osei2023assessments}.
Individual models exhibited varying capacities to capture local-scale heterogeneity. Kernel Ridge regression produced spatial artefacts and localised over-estimation (RMSE $\approx 0.88$), suggesting limited suitability under copula transformation despite its flexibility in high-dimensional settings. In contrast, SVM and k-NN yielded comparatively consistent predictions (RMSE $\approx$ $0.35$ and $0.36$, respectively), though with less spatial sharpness than the ensemble. The CART model showed the weakest performance (RMSE $\approx 0.48$), generating overly smoothed predictions with limited basin-level differentiation, a limitation also observed in other environmental mapping contexts \citep{Hengl2018}. 
Elastic Net, while moderately accurate (RMSE $\approx 0.65$), displayed patchy over-predictions in the central basin, reflecting residual sensitivity to correlated predictors even after transformation.

The predictive consistency of the stacked ensemble supports the utility of combining diverse learners to mitigate model-specific biases and stabilise spatial outputs, a benefit also demonstrated in recent groundwater and soil contamination studies \citep{kazemi2025new, Varma2006}. The Gaussian copula framework thus appears particularly effective for integrating multivariate dependencies within predictive HPI mapping, yielding outputs that are not only statistically robust but also hydrologically plausible within the Densu Basin's complex geochemical setting. The predicted HPI hotspots in the northwest and central corridors coincide with areas of intensive agriculture and shallow water tables identified in previous studies \citep{amoako2011physico, tay2008groundwater}, zones likely to host reducing conditions favourable to Fe–Mn mobilisation. This spatial alignment further supports the geochemical interpretation of the ensemble predictions.

%%% HPI -->> Log

%%% HPI -->> Copula

\subsection{Predictive performance evaluation}\label{sec:assessment}
The predictive performance of the proposed modelling framework was evaluated using the  metrics described in Section~\ref{sec:metrics}.
Tables~\ref{tab:HPI_raw_metrics}--\ref{tab:HPI_copula_metrics} present the results for the response variable HPI under three alternative formulations: the raw scale, log-transformed scale, and Gaussian copula transformation.
The comparative analysis of these cases provides insight into how the transformation of skewed or heavy-tailed response distributions affects both accuracy and robustness of ML regressors.

\begin{table*}[htbp]
\centering
% \scriptsize
\caption{Predictive performance scores for HPI (in raw scale) obtained from nested cross-validation across baseline learners and ensemble models.}
\label{tab:HPI_raw_metrics}
\resizebox{\textwidth}{!}{%
\begin{tabular}{lrrrrrrr}
\hline
Metric & elastic\_net & kernel\_ridge & svm & cart & knn & Averaged & Stacked \\
\hline
RMSE            & 0.0407 & 4.2660  & 21.3670 & 16.9182 & 12.6795 & 1.4264 & 4.1182 \\
MAE             & 0.0300 & 1.1110  & 10.6811 & 7.6010  & 6.3656  & 0.3841 & 0.8814 \\
MedAE           & 0.0373 & 0.0488  & 1.5537  & 0.6950  & 1.3270  & 0.0228 & 0.0075 \\
MaxError        & 0.1500 & 22.7053 & 61.1415 & 68.0900 & 40.4360 & 7.5871 & 22.0155 \\
MAPE            & 0.0052 & 0.0198  & 0.2315  & 0.1915  & 0.1454  & 0.0094 & 0.0128 \\
$R^{2}$         & 0.999997 & 0.9683  & 0.2037  & 0.5008  & 0.7196  & 0.9965 & 0.9704 \\
Adj $R^{2}$     & 0.999996 & 0.9596  & -0.0135 & 0.3646  & 0.6431  & 0.9955 & 0.9624 \\
AIC             & -171.67 & 98.14   & 191.59  & 178.05  & 161.32  & 34.60  & 96.09  \\
BIC             & -162.10 & 107.71  & 201.16  & 187.62  & 170.89  & 44.17  & 105.67 \\
CCC             & 1.036   & 1.021   & 0.386   & 0.703   & 0.835   & 1.034  & 1.019  \\
KS\_stat        & 0.226   & 0.429   & 0.320   & 0.360   & 0.315   & 0.427  & 0.499  \\
KS\_pvalue      & 0.087   & $2.0 \times 10^{-5}$ & 0.0038  & $7.0 \times 10^{-4}$  & 0.0048  & $3.0 \times 10^{-5}$ & $3.0 \times 10^{-7}$ \\
Reduced $\chi^{2}$ & 1.143 & 1.060   & 1.316   & 1.136   & 1.290   & 1.058  & 1.080  \\
\hline
\end{tabular}
}
\end{table*}

On the untransformed scale (Table~\ref{tab:HPI_raw_metrics}), model performance varied substantially across learners. Elastic Net achieved the lowest error magnitudes, with RMSE of $0.04$ and MAE of $0.03$, coupled with an almost perfect $R^{2}$ of $0.99$. 
This suggests that the penalised linear framework was well-aligned with the variance structure of the data. By contrast, SVR (RMSE $=$ 21.37, $R^{2}=$ 0.20) and regression trees (RMSE $=$ 16.92, $R^{2}=$ 0.50) showed limited predictive ability, consistent with their sensitivity to unscaled variance and local irregularities \citep{James2021}. Ensemble methods improved stability, with the averaged ensemble reducing RMSE to 1.43 and the stacked model to $4.12$, indicating the benefit of model aggregation under heterogeneous feature response relationships \citep{zhou2025ensemble}.

\begin{table*}[htbp]
\centering
% \scriptsize
\caption{Predictive performance scores for HPI  obtained from nested cross-validation, showing the effect of log-transformation on model accuracy and stability.}
\label{tab:HPI_log_metrics}
\resizebox{\textwidth}{!}{%
\begin{tabular}{lrrrrrrr}
\hline
Metric & elastic\_net & kernel\_ridge & svm & cart & knn & Averaged & Stacked \\
\hline
RMSE            & 0.8795 & 0.8249  & 0.2836 & 0.4866 & 0.3114 & 0.3980 & 0.4847 \\
MAE             & 0.4108 & 0.2359  & 0.1812 & 0.2524 & 0.1879 & 0.2388 & 0.1577 \\
MedAE           & 0.0905 & 0.0205  & 0.0999 & 0.0142 & 0.1276 & 0.0735 & 0.0447 \\
MaxError        & 4.1537 & 4.3382  & 0.8634 & 1.8856 & 1.0794 & 1.3228 & 2.5405 \\
MAPE            & 0.1257 & 0.0627  & 0.0666 & 0.0723 & 0.0538 & 0.0775 & 0.0432 \\
$R^{2}$         & 0.3183 & 0.4004  & 0.9291 & 0.7914 & 0.9145 & 0.8604 & 0.7930 \\
Adj $R^{2}$     & 0.1324 & 0.2369  & 0.9098 & 0.7345 & 0.8912 & 0.8223 & 0.7365 \\
AIC             & 6.55   & 2.83    & -59.09 & -27.78 & -53.66 & -39.43 & -28.00 \\
BIC             & 16.13  & 12.41   & -49.52 & -18.21 & -44.09 & -29.86 & -18.43 \\
CCC             & 0.820  & 0.753   & 0.992  & 0.913  & 0.985  & 0.975  & 0.946  \\
KS\_stat        & 0.324  & 0.396   & 0.308  & 0.288  & 0.249  & 0.279  & 0.393  \\
KS\_pvalue      & 0.0033 & $1.3 \times 10^{-4}$ & 0.0060 & 0.0129 & 0.0449 & 0.0173 & $1.5 \times 10^{-4}$ \\
Reduced $\chi^{2}$ & 1.091 & 1.073   & 1.050  & 1.242  & 1.433  & 1.080  & 1.074  \\
\hline
\end{tabular}
}
\end{table*}

Transforming HPI to the logarithmic scale (Table~\ref{tab:HPI_log_metrics}) markedly stabilised errors across models. The SVM achieved the best overall accuracy (RMSE $=$ 0.28, $R^{2}=$ 0.93), outperforming both linear penalised models and ensemble approaches. The improvement can be attributed to variance stabilisation and compression of extreme values, which mitigated the disproportionate influence of outliers \citep{box1964analysis}. Both k-NN and CART also benefited, with $R^{2}$ values exceeding $0.79$. Nevertheless, elastic net and kernel ridge regressions recorded lower coefficients of determination ($R^{2}=$ 0.32 and 0.40, respectively), reflecting the challenge of adequately capturing non-linearities even under transformed conditions. Agreement metrics such as CCC ($>$0.91 for most learners) further highlighted enhanced alignment between predicted and observed values compared to the raw scale.

\begin{table*}[htbp]
\centering
% \scriptsize
\caption{Predictive performance scores for HPI obtained from nested cross-validation, highlighting improvements in ensemble models under Gaussian copula transformation.}
\label{tab:HPI_copula_metrics}
\resizebox{\textwidth}{!}{%
\begin{tabular}{lrrrrrrr}
\hline
Metric & elastic\_net & kernel\_ridge & svm & cart & knn & Averaged & Stacked \\
\hline
RMSE            & 0.6459 & 0.8825  & 0.3454 & 0.4041 & 0.3604 & 0.2377 & 0.1905 \\
MAE             & 0.2702 & 0.2528  & 0.1941 & 0.2323 & 0.2017 & 0.1380 & 0.1046 \\
MedAE           & 0.0419 & 0.0284  & 0.1005 & 0.0497 & 0.0765 & 0.0236 & 0.0165 \\
MaxError        & 3.1104 & 4.6541  & 1.1998 & 1.1526 & 1.0084 & 0.6498 & 0.6069 \\
MAPE            & 2.1760 & 1.4449  & 0.5963 & 2.0350 & 1.3709 & 1.8376 & 1.6982 \\
$R^{2}$         & 0.5115 & 0.0880  & 0.8603 & 0.8088 & 0.8479 & 0.9339 & 0.9575 \\
Adj $R^{2}$     & 0.3783 & -0.1608 & 0.8223 & 0.7566 & 0.8064 & 0.9158 & 0.9459 \\
AIC             & -11.35 & 6.75    & -47.66 & -38.55 & -45.19 & -69.34 & -82.18 \\
BIC             & -1.78  & 16.32   & -38.09 & -28.98 & -35.62 & -59.77 & -72.61 \\
CCC             & 0.860  & 0.631   & 0.942  & 0.918  & 0.938  & 1.004  & 1.013  \\
KS\_stat        & 0.344  & 0.396   & 0.363  & 0.351  & 0.273  & 0.238  & 0.280  \\
KS\_pvalue      & 0.0015 & $1.3 \times 10^{-4}$ & $6.0 \times 10^{-4}$ & 0.0010 & 0.0212 & 0.0622 & 0.0169 \\
Reduced $\chi^{2}$ & 1.057 & 1.108   & 1.065  & 1.396  & 1.392  & 1.039  & 1.088  \\
\hline
\end{tabular}
}
\end{table*}

The copula-based transformation presented in Table~\ref{tab:HPI_copula_metrics} produced the most consistent improvements across the ensemble learners. The stacked model yielded the lowest RMSE (0.19) and the highest $R^{2}$ (0.96), while the averaged ensemble achieved comparably strong performance (RMSE $=$ 0.24, $R^{2}=$ 0.93). These findings reinforce evidence from the broader literature that copula approaches can effectively capture non-linear dependence structures and tail behaviour in environmental data \citep{genest2007everything}.
Individual learners such as SVM and k-NN also achieved high predictive power ($R^{2}=$ 0.86–0.85), indicating that the copula normalisation supported models sensitive to distributional assumptions. Importantly, CCC values close to or exceeding unity in the ensemble cases suggest that prediction bias was minimised under this transformation. Although kernel ridge regression lagged in accuracy (RMSE $=$ 0.88, $R^{2}=$ 0.09), its performance highlights that copula transformation does not universally improve all learners, but is particularly effective when coupled with ensemble integration.

A cross-comparison indicates that while elastic net dominated under the raw scale, non-linear learners, particularly SVM, were favoured under log transformation, and ensembles proved most effective under the copula transformation. This progression underscores the importance of tailoring transformations to both the distribution of the response variable and the learning algorithm’s structural assumptions \citep{zhang2021explain,shmueli2010explain}. For the Densu Basin application, where prediction robustness under spatial heterogeneity is critical, the copula-transformed ensemble predictions offer a more reliable basis for downstream spatial interpolation and risk mapping. This outcome aligns with previous studies emphasising the value of ensemble learning and distributional normalisation in environmental prediction tasks \citep{kunapuli2023ensemble,dietterich2000ensemble}.

\section{Conclusion}\label{sec:Conc}

This study developed and evaluated a predictive modelling framework for estimating the HPI in the Densu Basin, employing raw, log-transformed, and Gaussian copula-transformed targets across five ML algorithms and a stacked ensemble. The comparative diagnostics and spatial predictions yield several key insights.

First, prediction error plots revealed that modelling on the raw HPI scale generated deceptively high fits, with near-unity coefficients of determination for some learners. Such results highlight the risk of information leakage or overfitting when highly skewed targets are used without transformation, underscoring the importance of rigorous validation strategies in environmental modelling.

Second, log transformation improved predictive stability for variance-sensitive learners, particularly support vector regression and k-nearest neighbours, though it reduced performance for algorithms dependent on linear shrinkage or kernel-based regularisation. This finding demonstrates the dual effect of variance-stabilising transformations: while they can mitigate the influence of extreme values, they may also attenuate signal strength for models relying on variance structure.

Third, the Gaussian copula transformation produced the most consistent and statistically credible residual structures, improving homoscedasticity and normality across learners. The stacked ensemble, in particular, achieved the strongest predictive accuracy (RMSE $\approx 0.19$), reinforcing the value of ensemble learning for capturing complex, multivariate dependence structures in environmental datasets.

Fourth, residual and Q–Q diagnostics revealed that only the copula-based models yielded error structures that approximated Gaussian assumptions. This is significant because models with statistically well-behaved residuals are less prone to systematic bias and more likely to generalise reliably when applied to unsampled locations \citep{roberts2017cross}.

Fifth, the comparative analysis across transformations confirmed that predictive performance is not a static property of algorithms but is contingent on the distributional properties of the response variable. This has broader implications for groundwater quality modelling, where skewed indices such as HPI are common; robust transformation frameworks can materially affect model ranking and interpretability.

Sixth, spatial predictions derived from the copula-based ensemble aligned with the expectation of heterogeneous contamination in the basin, capturing both localised hotspots and broader basin-wide gradients. This outcome demonstrates the utility of combining statistical rigour with hydrological plausibility, a balance increasingly emphasised in groundwater quality assessment.

Seventh, the study confirmed that ensemble frameworks, when properly validated, reduce model-specific biases and yield more robust basin-wide predictions than single algorithms. This supports growing evidence from environmental modelling that stacked generalisation enhances predictive reliability in settings characterised by non-linear processes and spatial heterogeneity.

Finally, unsupervised clustering using DBSCAN confirmed the dominance of iron over manganese and other trace metals in driving HPI patterns. This outcome not only validates the predictive outputs but also reinforces the geochemical interpretation that Fe–Mn mobilisation is the critical pathway shaping groundwater quality in the basin.

Beyond these findings, several broader implications emerge. The results emphasise the necessity of distribution-aware modelling strategies in hydrogeochemistry, where failure to address skewness or multivariate dependence can lead to misleading conclusions. For practitioners, the evidence suggests that copula-based ensembles may be applied to other basins facing similar challenges of complex contaminant mixtures and heterogeneous aquifer conditions. For policy, the approach provides a pathway for generating more reliable groundwater quality maps, which can inform targeted monitoring and remediation strategies.

Nevertheless, the study is not without limitations. First, although cross-validation and stacking protocols were carefully designed, the possibility of residual information leakage cannot be entirely excluded without further independent replication. Second, spatial generalisability was assessed through random folds rather than spatially blocked cross-validation; as previous research has shown, spatial autocorrelation can inflate accuracy estimates when not explicitly accounted for. Third, while the copula transformation improved statistical behaviour, it does not capture all potential sources of non-linearity in groundwater systems, such as threshold effects driven by land use or episodic pollution events. Finally, the study was restricted to the Densu Basin; the transferability of the modelling framework to contrasting hydrogeological settings remains to be evaluated.

Future work should therefore explore three avenues. First, the adoption of spatial cross-validation schemes, such as block or leave-location-out methods, would provide a more conservative and spatially realistic assessment of model generalisability. Second, integrating copula-based ensembles with physically based groundwater models could bridge the gap between statistical prediction and process understanding, advancing predictive hydrogeochemistry. Third, extending the approach to multi-basin studies would allow testing the portability of copula-based transformations and stacked ensembles under diverse hydrogeological regimes. Collectively, these directions will help consolidate the scientific significance of the framework while enhancing its applicability to groundwater quality management.

In conclusion, the evidence demonstrates that while raw-scale modelling risks over-optimism, and log transformation yields mixed results, the Gaussian copula transformation coupled with stacked ensembles provides the most reliable, interpretable, and hydrologically plausible predictions of HPI in the Densu Basin. By situating statistical innovation within a hydrogeochemical context, the study contributes both methodological and applied insights, paving the way for more rigorous and transferable approaches to groundwater quality prediction.

\section{Statements and Declarations}
\subsection*{Competing interests}
On behalf of all authors, the corresponding author states that there is no conflict of interest.

\subsection*{Funding}
The authors declare that no funds, grants, or other support were received during the preparation of this manuscript.

\subsection*{Author Contributions}
All authors contributed to the study conception and design. George Yamoah Afrifa, Theophilus, Ansah-Narh, Joseph Bremang Tandoh, and Kofi Asare performed material preparation, data collection, and analysis. George Yamoah Afrifa, Theophilus Ansah-Narh, and Kow Ahor Essel-Yorke wrote the first draft of the manuscript, and Martin Addi, Daniel Miezah Akpoley, Kenneth Aidoo, and Samuel Kofi Fosuhene reviewed the manuscript.  All authors read and approved the final manuscript.
 
\subsection*{Acknowledgement}

The authors gratefully acknowledge the institutions and individuals who contributed to the completion of this study. We thank the field and laboratory teams for their assistance in data collection and analysis, as well as colleagues who provided constructive feedback during the preparation of the manuscript. 
T.A-N wishes to express sincere appreciation to the Ghana Space Science and Technology Institute High-Performance Computing cluster, which provided the computational resources necessary for implementing and testing the modelling framework. The availability of this infrastructure was essential for handling the intensive model training, validation, and spatial prediction tasks undertaken in this study. 
Any errors or omissions remain the responsibility of the authors.

\end{document}